\documentclass[conference]{IEEEtran}
\usepackage{cite}
\usepackage{amsmath,amssymb,amsfonts}
\usepackage{algorithmic}
\usepackage{graphicx}
\usepackage{textcomp}
\usepackage{xcolor}
\usepackage{fancyhdr}
\usepackage{authblk}
\usepackage[hyphens]{url}
\usepackage[belowskip=-10pt,aboveskip=3pt]{caption}

\usepackage{array, makecell}
\usepackage[ruled, vlined]{algorithm2e} 

\SetCommentSty{mycommfont}
\def\BibTeX{{\rm B\kern-.05em{\sc i\kern-.025em b}\kern-.08em
    T\kern-.1667em\lower.7ex\hbox{E}\kern-.125emX}}
\usepackage[customcolors]{hf-tikz}
\usepackage[bookmarks=true,breaklinks=true,letterpaper=true,colorlinks,linkcolor=black,citecolor=blue,urlcolor=black]{hyperref}

\def\BibTeX{{\rm B\kern-.05em{\sc i\kern-.025em b}\kern-.08em
    T\kern-.1667em\lower.7ex\hbox{E}\kern-.125emX}}

\title{VersaGNN: a \underline{Versa}tile accelerator for \underline{G}raph \underline{N}eural \underline{N}etworks} 
\author[1]{Feng Shi}
\author[1]{Ahren Yiqiao Jin}
\author[1]{Song-Chun Zhu}
\affil[1]{University of California Los Angeles}

\begin{document}
\maketitle
\pagestyle{plain}


\begin{abstract}
\textit{Graph Neural Network} (GNN) is a promising approach for analyzing graph-structured data that tactfully captures their dependency information via node-level message passing. It has achieved state-of-the-art performances in many tasks, such as node classification, graph matching, clustering, and graph generation. As GNNs operate on non-Euclidean data, their irregular data access patterns cause considerable computational costs and overhead on conventional architectures, such as GPU and CPU. Our analysis show that GNN adopts a hybrid computing mode. The \textit{Aggregation} (or \textit{Message Passing}) phase performs vector additions where vectors are fetched with irregular strides. The \textit{Transformation} (or \textit{Node Embedding}) phase can be either dense or sparse-dense matrix multiplication. In this work, We propose \textit{VersaGNN}, an ultra-efficient, systolic-array based versatile hardware accelerator that unifies dense and sparse matrix multiplication. By applying this single optimized systolic-arrays to both aggregation and transformation phases, we have significantly reduced chip sizes and energy consumption. We then divide the computing engine into blocked systolic arrays to support the \textit{Strassen}'s algorithm for dense matrix multiplication, dramatically scaling down the number of multiplications and enabling high-throughput computation of GNNs. To balance the workload of sparse-dense matrix multiplication, we also introduced a greedy algorithm to combine sparse sub-matrices of compressed format into condensed ones to reduce computational cycles. Compared with current state-of-the-art GNN software frameworks, \textit{VersaGNN} achieves on average 3712$\times$ speedup with 1301.25$\times$ energy reduction on CPU, and 35.4$\times$ speedup with 17.66$\times$ energy reduction on GPU.

\end{abstract}

\section{Introduction}
Graph Neural Networks (GNNs) have achieved state-of-the-art performances in node classification \cite{gcn-kipf2017,graphsage_2017}, link prediction \cite{kipf2016variational, link-pred-zhang}, graph classification \cite{graph-class-rex-ying}, graph generation \cite{kipf2016variational, Wang2017GraphGANGR}, and clustering \cite{wang-mgae-2017, graph-class-rex-ying} on arbitrarily structured graphs. The power of representation learning on graphs comes from feature embedding, which includes extracting structured, low dimensional features from unstructured, high dimensional graphs.

On one hand, traditional \textit{Convolutional Neural Networks} (CNNs) \cite{lecun1998gradient} that operate on Euclidean data are characterized by local connections and shared weights, and are able to extract multi-scale localized spatial features. Euclidean data, such as images, can be represented as regular grids in the Euclidean space. Thus, a CNN is able to exploit the shift-invariance and local connectivity of Euclidean data. On the other hand, GNNs inherit the irregular computing patterns and processing dataflow of graph analytics, resulting in the inefficient use of CPUs and GPUs. 

Meanwhile, the majority of real-word graphs for GNNs follow the Power-Law distribution - the number of nodes with degree $N$ is proportional to $N^\alpha$ for some constant $\alpha$ \cite{geng2020awbgcn}. Thus, a minority of nodes share high degrees, leading to remarkably unbalanced adjacency matrices. Therefore, hardware architectures must adapt to the varying shapes of the high-dimensional convolutions in GNN. Generally, two primary execution phases, \textbf{Aggregation} and \textbf{Transformation}, occupy the most execution time \cite{torch_geometric, graphsage_2017, gnn-survey2, hygcn}.

\textbf{Aggregation (\textbf{Message Passing}) Phase}: 
GNN follows a neighborhood aggregation strategy in which each node's representation is updated by aggregating features of its adjacencies. Common aggregation strategies include \textbf{Sum}, \textbf{Average}, \textbf{Mean}, and \textbf{Max}. After $N$ iterations of aggregation, a node’s representation captures the structural information within its $N$-hop network neighborhood. The main computation kernels are data loading, which collects feature vectors that are indexed by the neighboring node's addresses, and the execution of aggregation. Due to their sparsity, the feature vectors can be efficiently fetched by coalescing their elements. Within the feature matrix, the feature vectors of adjacent nodes can be separated by strides. Poor use of these inter-vertex data parallelism can incur significant cache misses and address calculations.

\textbf{Transformation (\textbf{Node Encoding}) Phase}: This phase is usually expressed as a Multi-layer Perceptron (MLP) that transforms node feature vectors to lower dimensional embeddings using Matrix-Vector Multiplication (MVM). The matrix multiplication can be either dense or sparse, depending on the sparsity of feature matrices at different convolutional layers. A non-linear activation function
is applied to each vertex to yield the outputs. This phase is characterized by regular computational graph and homogeneous data access patterns.

To accelerate GNN-based applications that harness these two distinct stages and process highly variable real-world graphs, we propose several optimization approaches for accelerating GNN with a software/hardware co-design paradigm. Our main contributions are:
\begin{itemize}
    \item We unify the processing of Aggregation and Transformation stages using a single processor that is competent in handling the irregular data access patterns and the hybrid computing mode of GNNs.
    \item We apply the classic Strassen's algorithm to accelerating dense matrix multiplication, significantly reducing the number of costly multiplications. 
    \item We exploit an ultra-efficient, greedy-based load-balancing approach that achieves considerable speedup in sparse-dense multiplication.
    \item We propose \textit{VersaGNN}, a high-throughput and memory-efficient Graph Neural Network accelerator based on the well-known systolic array design.
    \item We implement our architecture design using Chisel HDL \cite{bachrach2012chisel}, and test our \textit{VersaGNN} using four well-known GNN models on six benchmark graph datasets. Compared to the state-of-the-art software frameworks, our work achieves on average 3712$\times$ speedup with 1301.25$\times$ energy reduction on CPU, and 35.4$\times$ speedup with 17.66$\times$ energy reduction on GPU, respectively. 
\end{itemize}

\section{Background}
In this section, we review the core concepts of Graph Neural Networks. Table \ref{tab:notations} lists the notations of GNNs used throughout the paper.
\begin{table}[h]
\caption{Notations of GNN}
\resizebox{0.48\textwidth}{!}{%
\begin{tabular}{ll|ll}
\hline
\multicolumn{1}{c}{\textit{\textbf{Notation}}}   & \multicolumn{1}{c|}{\textit{\textbf{Description}}} & \multicolumn{1}{c}{\textit{\textbf{Notation}}}   & \multicolumn{1}{c}{\textit{\textbf{Description}}} \\ \hline
$\mathbf{G}$ & graph $\mathbf{G} = (V, E, H_0)$  & $V$            & nodes/vertices of $G$ \\ 
$E$          & edges of G                        & deg($v$)       & degree of node v \\
$e_{i,j}$    & edge between node i and j         & \textit{N(v)}  & neighbors of node v \\
$\Theta(\cdot)$          & Neural Networks                  & $a_v^{(i)}$    & aggregated feature vector of $v$ at layer $i$ \\
$A,\,A_{i,j}$& adjacency matrix, element at $(i, j)$ & $W^{(i)}$      & weight matrix of $i^{th}$ layer  \\
$h_v^{(i)}$  r    & hidden feature vector of node $v$ & $b^{(i)}$      & bias of $i^{th}$ layer \\
$H_0$        & initial state of \textit{feature matrix} & $\sigma(.)$ & activation function
\end{tabular}
}
\label{tab:notations}
\end{table}

\subsection{Graph Convolutions}
\begin{figure}[h]
    \centering
    \includegraphics[width=0.47\textwidth]{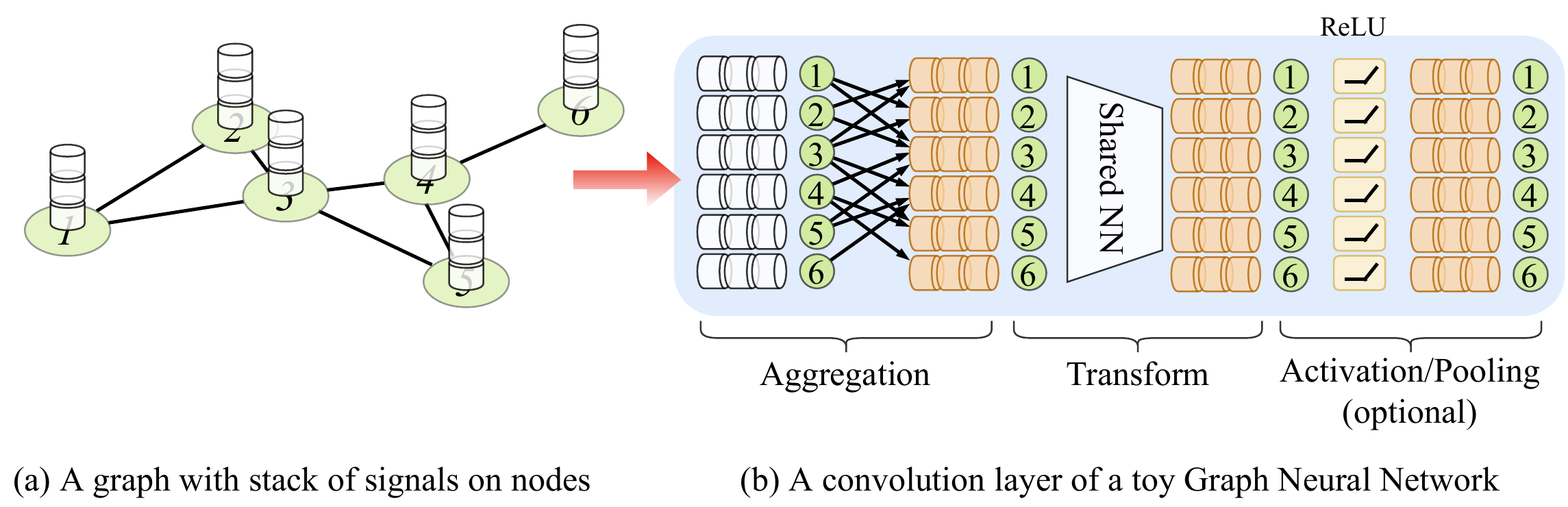}
    \caption{For the graph in (a), (b) shows how the node signals are passing through the graph convolution operator}
    \label{fig:toy_gcn}
\end{figure}

\textit{Graph Neural Networks} utilize the \textit{message passing} mechanism \cite{gnn-survey1, gnn-survey2} for graph node embedding, usually generated by \textit{graph convolutions}. A classic GNN is constructed with a stack of two or three graph convolution layers, whose structure is illustrated in Fig.\ref{fig:toy_gcn}. A graph convolution layer takes the feature matrix as the input, arranged by graph node signals, and performs convolution on the feature matrix, followed by one or two optional nonlinear operators, such as nonlinear activation (e.g ReLU, LeakyReLU) and pooling.

Convolutional GNNs can be categorized as spectral-based GNNs or spatial-based GNNs \cite{gnn-survey2}. Spectral-based approaches such as \cite{gcn-kipf2017, henaff2015deep, defferrard2016convolutional} define graph convolutions by adopting filters from the perspective of graph signal processing \cite{shuman2013emerging}. The graph convolution operations are interpreted as removing noises from graph signals. Spatial-based approaches, including \cite{atwood2016diffusion, gilmer2017neural, niepert2016learning}, exploits the information propagation paradigm and aims at collecting features of each node from its K-hop neighbors.

The \textit{convolution operator} in a graph convolution layer consists of two consecutive phases, the \textbf{Aggregation} (or \textbf{Message Passing}) phase, and the \textbf{Transformation} (or \textbf{Node Encoding}) phase. Equation (\ref{eq:gnn_msg}) gives a mathematical formulation of the message-passing network \cite{torch_geometric},
\begin{equation}
    \mathbf{h}_i^{(k)} =  \Theta_{i}^{(k)} \left( \mathbf{h}_i^{(k-1)}, \Xi_{j \in N(i)} \, \Theta_{j}^{(k)}\left(\mathbf{h}_i^{(k-1)}, \mathbf{h}_j^{(k-1)},\mathbf{e}_{j,i}\right) \right)
    \label{eq:gnn_msg}
\end{equation}
, where $\Xi$ embodies a differentiable and permutation-invariant aggregation function, e.g., \textbf{Sum}, \textbf{Mean}, or \textbf{Max}. $\Theta_{i}$ and $\Theta_{j}$ indicate neural networks or linear transformations. In the following subsections, we discuss some state-of-the-art models and their aggregation strategies.

\subsection{Direct Aggregation}
\textbf{GraphSAGE} \cite{graphsage_2017} exploits vertex features such as text attributes and node degrees to learn an embedding function, formulated as:
\begin{equation}
    \begin{split}
        \hat{\mathbf{h}}_i &= \Theta \left( \,\,\,
        \tikzmarkin{mean}\mathbf{mean}_{j \in N(i) \cup \{ i \}} \tikzmarkend{mean}\,\,\,(\mathbf{h}_j) \right)\\
    \mathbf{h}^{\prime}_i &= \frac{\hat{\mathbf{h}}_i}
        {\|\, \hat{\mathbf{h}}_i \, \|_2}
    \end{split}
    \label{eq:sage}
\end{equation}
where the feature aggregation is the \textbf{$mean$} operator marked in Equation (\ref{eq:sage}). Instead of training individual node-wise embedding vectors, \textbf{GraphSAGE} generates embeddings through uniform sampling and feature aggregation from the node's neighbors, thus taking into account both the node's own features as well as its neighboring features, and balance execution overhead with accuracy.

\textbf{GIN} \cite{gin_xu2018} utilizes the Graph Isomorphism Operator to enhance the representation power of GNNs. For each graph node, \textbf{GIN} recursively aggregates and transforms representation vectors of its adjacent nodes. With its high expressive power, \textbf{GIN} is able to capture the structural information of both large and small graphs.
\begin{equation}
    \mathbf{h}^{\prime}_i = \Theta \left( (1 + \epsilon) \cdot
        \mathbf{h}_i + \,\,
        \tikzmarkin{gin}(0.05,-0.5)(-0.1,0.5)
        \sum_{j \in N(i)} 
        \tikzmarkend{gin}\,\,\mathbf{h}_j
        \,\,\right)
    \label{eq:gin}
\end{equation}
$\epsilon$ is a learnable parameter that improves the node's self-confidence, and $\Theta$ represents a Multi-Layer Perceptron (MLP). \textbf{GIN} uses summation as its aggregation operator, which can represent universal functions over multisets, as marked in Equation (\ref{eq:gin}).

\subsection{Weighted Aggregation}

\textbf{GCN} \cite{gcn-kipf2017} is composed of two or three convolutional layers with residual connections. 
\begin{equation}
\mathbf{H}^{\prime} = \mathbf{\hat{D}}^{-1/2} \mathbf{\hat{A}}
        \mathbf{\hat{D}}^{-1/2} \mathbf{H}\Theta
    \label{eq:gcn}
\end{equation}
$\mathbf{\hat{A}} = \mathbf{A} + \mathbf{I}$ denotes the adjacency matrix with self-loops, and $\hat{D}_{i,i} = \sum \hat{A}_{i,j}$ represents its diagonal degree matrix. Expanding Equation (\ref{eq:gcn}), we get
\begin{equation}
    \mathbf{h}_i^{(k)} = \Theta \left( 
    \sum_{ j \in N(i) \cup \{ i \} }  
    \tikzmarkin{gcn}(0.1,-0.5)(-0.03,0.6)
    \dfrac{1}{\sqrt{\deg(i)} \cdot \sqrt{deg(j)}} 
    \tikzmarkend{gcn} 
    \,\,\,\cdot \mathbf{h}_j^{(k-1)} \right)
    \label{eq:gcn_expand}
\end{equation}
We observe from Equation (\ref{eq:gcn_expand}) that the feature aggregation phase performs a weighted summation (aggregation) of neighbor features. The coefficients within the marked region are derived from the degree matrix and perform a degree-based normalization.

\textbf{GAT} \cite{gat_2018} leverages masked self-attention layers in feature aggregation to assign computation importance to each nodes. It does not require costly matrix operations as well as pre-knowledge of graph structural information, and is able to achieve computation efficiency including intra-graph parallelization. 
\begin{equation}
    \begin{split}
        \mathbf{h}^{\prime}_i &= \,\,
        \tikzmarkin{gat1}
        \alpha_{i,i}
        \tikzmarkend{gat1}
        \,\,\,\Theta\mathbf{h}_{i} +
        \sum_{j \in N(i)} 
        \tikzmarkin{gat2}
        \alpha_{i,j}
        \tikzmarkend{gat2}
        \,\,\,\Theta\mathbf{h}_{j} \\
         &= \Theta \left( \sum_{j \in N(i) \cup \{ i\} } 
         \tikzmarkin{gat3}
         \alpha_{i, j} 
         \tikzmarkend{gat3}
         \cdot \mathbf{h}_{j} \right)
    \end{split}
    \label{eq:gat1}
\end{equation}
The attention coefficient $\alpha_{i,j}$ denoting the importance of node $j$’s features to node $i$ is computed as
\begin{equation}
    \alpha_{i,j} = \textit{SoftMax} \left( \textit{LeakyReLU} \left(\mathbf{a}^{\top}
        [\Theta\mathbf{h}_i \, \Vert \, \Theta\mathbf{h}_j] \right) \right)
    \label{eq:gat2}
\end{equation}
Equation (\ref{eq:gat1}) uses attention coefficients for the weighted aggregation of neighbor node embeddings. The denominator marked in Equation (\ref{eq:gat2}) also has a form similar to aggregation, where \textit{SoftMax} is a normalization performed after the node-level exponential operation $exp(\cdot)$.

\section{Motivation}
GNNs can be considered as an extension of classic deep neural networks with irregular topology that support graph-structured inputs and outputs. Specialized architecture designs for GNNs are required because existing machine learning accelerators are not suited to GNNs for the following reasons.

\textbf{Compounded Execution Mode}
Most GNN models are shallow with approximately three layers. The weight matrices $W$ are generally dense as the result of layer-wise node aggregation. Current sparse matrix accelerators are specialized solely in processing sparse data formats, but will suffer significant overhead with dense matrix operations. However, the execution flow of GNNs follows a hybrid mode. Despite the sparsity of the input node feature matrices, the following aggregation operations will gradually populate the intermediary node embeddings with non-zero values. On the other hand, node feature transformation involves dense matrix multiplications. Present GNN accelerators are designed either to undertake the sparsity in \textbf{Aggregation}, or to employ the regularity in dense matrix multiplications that constitute \textbf{Transformation}, but lack the generality to handle both cases.

\textbf{Workload Imbalance}
The irregular access and computation patterns of the \textbf{Aggregation} phase, which involves graph traversals that require tremendous memory access relative to only small amounts of calculation, make the mainstream computation platforms unsuitable for GNNs. CPUs and GPUs are not capable of irregular data movements and computations that constitute GNN operations. Their inefficiency in memory access causes the waste of off-chip memory bandwidths. Although modern GPUs such as NVIDIA A100 support sparse matrix multiplication with pruning techniques; for many real-world graph datasets, their feature matrices cannot be pruned, as the features are represented using binary values. Meanwhile, the adjacency matrices are subject to significant loss of graph structural information, making compression unrealistic despite their sparsity.

For traditional computation platforms including GPUs, their performance bottleneck on GNNs originates from their inability to settle the irregularity in Aggregation phase. Their relatively high performance on GNNs is mostly attributed to the high-bandwidth memory, which incurs considerable energy consumption. Furthermore, although they leverage the regularity in \textbf{Transformation} phase, the data copying and synchronization between threads for parameter reusing are expensive. Meanwhile, graph analytics accelerators are only optimized to alleviate irregularity or exploit regularity.

\textbf{Usecase Generality} 
Aggregation can be direct aggregation or weighted aggregation. The weighted one is not just performing summation or other aggregation operation. Neighbor node's feature vector is scaled with a factor before applying the aggregation operator, e.g., Equation \ref{eq:gat1} of \textit{GAT}. Current GNN accelerators such as \textit{EnGN} \cite{he2020engn} and \textit{HyGCN} \cite{hygcn} are only optimized for the directed aggregation case. This scaling issue can be solved by sparse-dense matrix multiplication and is more efficient than other designs.

\textit{HyGCN} \cite{hygcn} is a GNN accelerator with a hybrid architecture. To harness the hybrid execution patterns of GNNs, \textit{HyGCN} separates the modules for the regular neural network processing and irregular graph processing. The \textit{Aggregation Engine} and \textit{Combination Engine}, used for node aggregation and transformation respectively, each require separate on-chip buffers which consume considerable chip area. Despite their potentials for pipelining, the distinct computing patterns of these two processing stages make it difficult to harness both modules for efficient processing of general GNN structures.

\textit{EnGN} \cite{he2020engn} adopts the Ring-Edge-Reduce(RER) dataflow to tames the poor locality of sparsely connected vertices, and manipulates the ring-edge-reduce (RER) PE-array to practice RER dataflow. However, RER's storage of neighbor node indices consumes  enormous local registers. Moreover, the index comparisons incurred by RER data movements may induce considerable latency and soaring computation cycles.

\begin{figure}[h]
    \centering
    \includegraphics[width=0.30\textwidth]{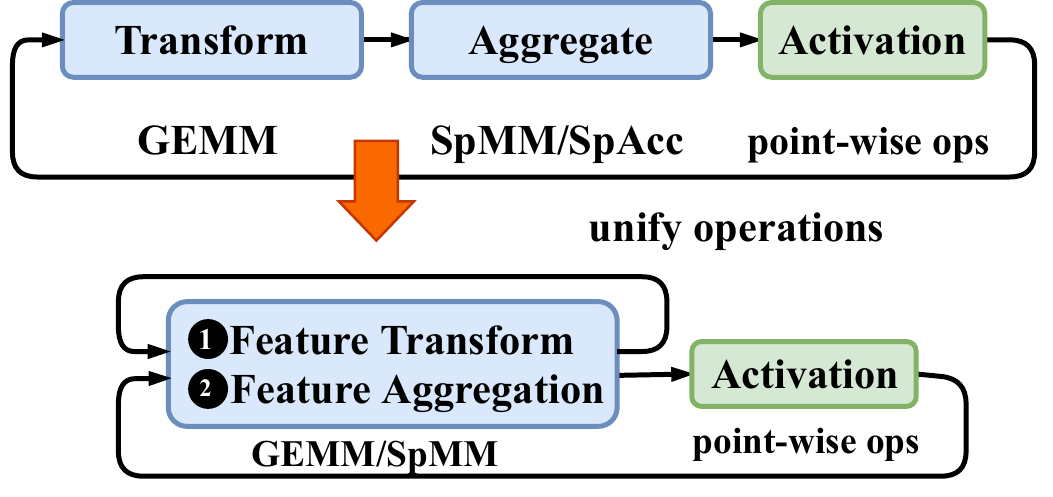}
    \caption{Unify transformation and aggregation phases}
    \label{fig:unify}
\end{figure}

The characteristics of \textbf{Aggregation} phase reveals that it can adopt sparse matrix multiplication, then it provides an option to unify the \textit{Transformation} phase and \textit{Aggregation} phase into one single arithmetic operation, i.e., matrix multiplication, but aggregation phase tends to be sparse-dense case. Fig.\ref{fig:unify} demonstrates the possibility of unification.

\textbf{Our Solution}
Therefore, we propose to accelerate the two phases with one versatile accelerator. We propose a unified hardware design that can reuse the limited on-chip buffers among the different processing stages. We can turn the aggregation stage into pseudo sparse matrix multiplication. Our design is a scalable and efficient parallel processing engine and support large-scale GNNs.

\section{Microarchitecture}
\begin{figure}
    \centering
    \includegraphics[width=0.45\textwidth]{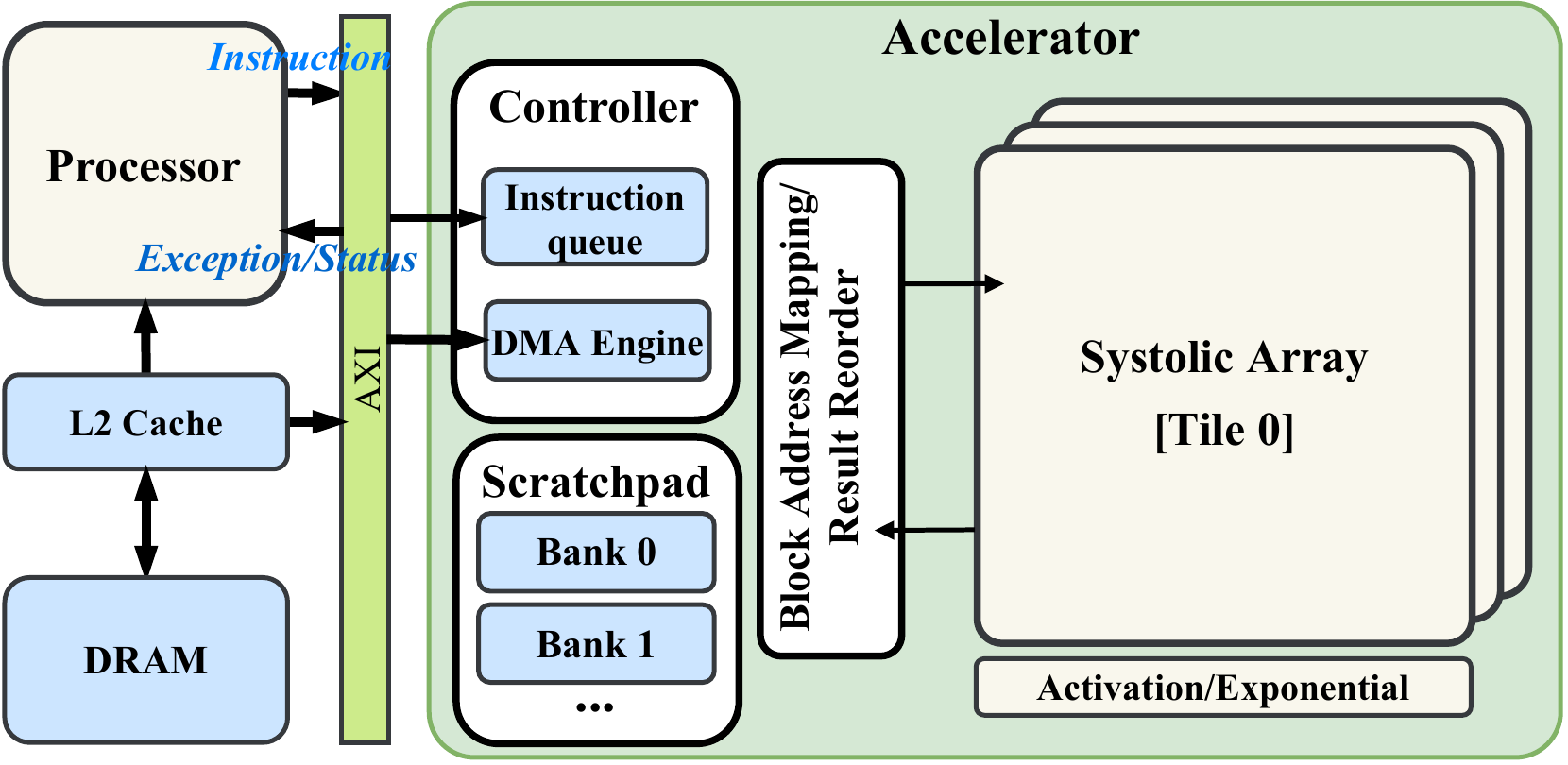}
    \caption{Overview of the hardware system design}
    \label{fig:chip}
\end{figure}
In this section, we discuss the hardware design of \textit{VersaGNN}, with its system-level overview outlined in Fig.\ref{fig:chip}. \textit{VersaGNN} consists of a general-purpose processor, a memory management interface, a multi-purpose bus, and an accelerator. The processor is used to control the whole system, send instructions to both the accelerator and the memory system, and collect the status of the accelerator. The accelerator comprises an instruction queue, a DMA engine, a scratchpad memory with several banks, and the systolic array tiles which are interconnected side-by-side in a chain/ring fashion. An additional \textit{Block Address Mapping} module is equipped for address matching when tiling is applied, and a {Result Reordering} module is used for the write-back phase of \textit{sparse-dense matrix multiplications} (SpMM). Since the bulk of GNN calculations are matrix computations, our accelerator targets the acceleration of both dense and sparse matrix addition and multiplication. 

\subsection{Systolic Style Matrix Multiplier}
The \textit{Transformation} phase of the Graph Neural Network, formed by Multi-Layer Perceptrons (MLP), is essentially a multiplication of node feature matrix with layer weight matrix \cite{gnn-survey1, gnn-survey2} that maps the high dimensional node features to lower dimensional spaces. The input feature vector of a graph node is usually very sparse \cite{gnn-survey2}, implying that a node does not hold all its defined features, with absent features holding zero values. The intermediary node embeddings are gradually populated with non-zero values as feature aggregation proceeds, whereas most weight matrices for these layers are dense. To fuse matrix operations among these disparate structures, GNN requires a versatile accelerator architecture to tackle both dense and sparse matrix calculations.

\subsubsection{Revisiting \textbf{Strassen's Algorithm}}
Before diving into systolic array tile design, we revisit \textit{Strassen's algorithm} \cite{wiki-strassen}, and expand it to block-based dense matrix multiplications to reduce the volume of expensive arithmetic multiplications.

Given two matrices (or matrix tiles) $A$ and $B$ as input, we partition each of them into 4 sub-blocks: $A_0, \dots, A_3$ for $A$, and $B_0, \dots, B_3$ for $B$. Similarly, the result matrix $C$ can be divided into $C_0, \dots, C_3$. 
{\smaller
    \begin{equation}
        \centering
        \begin{split}
        \tikzmarkin{a}
            S_4 & = B_2 - B_0  \\
            S_0 & = A_3 + A_0 \quad S_1 = B_0 + B_3 \\
            S_6 & = A_2 - A_0 \quad S_7 = B_0 + B_1 
        \tikzmarkend{a} \\
        \tikzmarkin[set fill color=green!50!lime!30,
        ]{b}
            M_0 & = S_0 \times S_1 \quad C_0 \mathrel{{+}{=}} M_0  \\ 
            M_3 & = A_3 \times S_4 \quad C_2 \mathrel{{+}{=}} M_3 \\ 
            M_5 & = S_6 \times S_7 \quad C_3 \mathrel{{+}{=}} M_5 
        \tikzmarkend{b} \\
        \tikzmarkin[set fill color=orange!50!lime!30,
        ]{c}
            C_0 & \mathrel{{+}{=}} M_3  \\
            C_3 & \mathrel{{+}{=}} M_0 
        \tikzmarkend{c} \\
        \end{split}
        \quad
        \begin{split}
        \tikzmarkin{d}
            S_2 & = A_2 + A_3  \\
            S_8 & = A_1 - A_3 \quad S_9 = B_2 + B_3  \\
            S_5 & = A_1 + A_0 \quad S_3 = B_1 - B_3 
        \tikzmarkend{d} \\
        \tikzmarkin[set fill color=green!50!lime!30,
        ]{e}
            M_1 & = S_2 \times B_0 \quad C_2 \mathrel{{+}{=}} M_1  \\ 
            M_2 & = A_0 \times S_3 \quad C_3 \mathrel{{+}{=}} M_2  \\
            M_4 & = S_5 \times B_3 \quad C_1 \mathrel{{+}{=}} M_4  \\
            M_6 & = S_8 \times S_9 \quad C_0 \mathrel{{+}{=}} M_6 
        \tikzmarkend{e} \\
        \tikzmarkin[set fill color=orange!50!lime!30,
        ]{f}
            C_3 & \mathrel{{-}{=}} M_1  \\
            C_1 & \mathrel{{+}{=}} M_2  \\
            C_0 & \mathrel{{-}{=}} M_4 
        \tikzmarkend{f}
        \end{split}
        \label{eq:strassen_algo}
    \end{equation}
}
\begin{center}
    \begin{figure*}[h]
        \centering
        \includegraphics[width=0.9\textwidth]{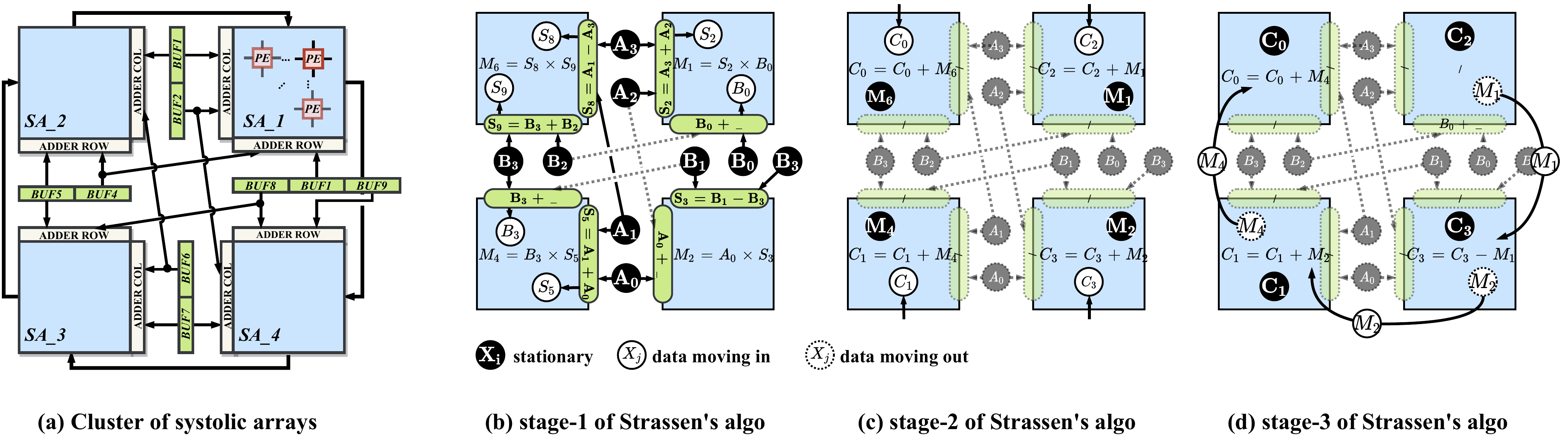}
        \caption{Scenario of \textbf{Strassen's Algorithm}: \textit{dataflow} and \textit{processing}}
        \label{fig:strassen_scenario}
    \end{figure*}
\end{center}

\begin{figure}
    \centering
    \includegraphics[width=0.47\textwidth]{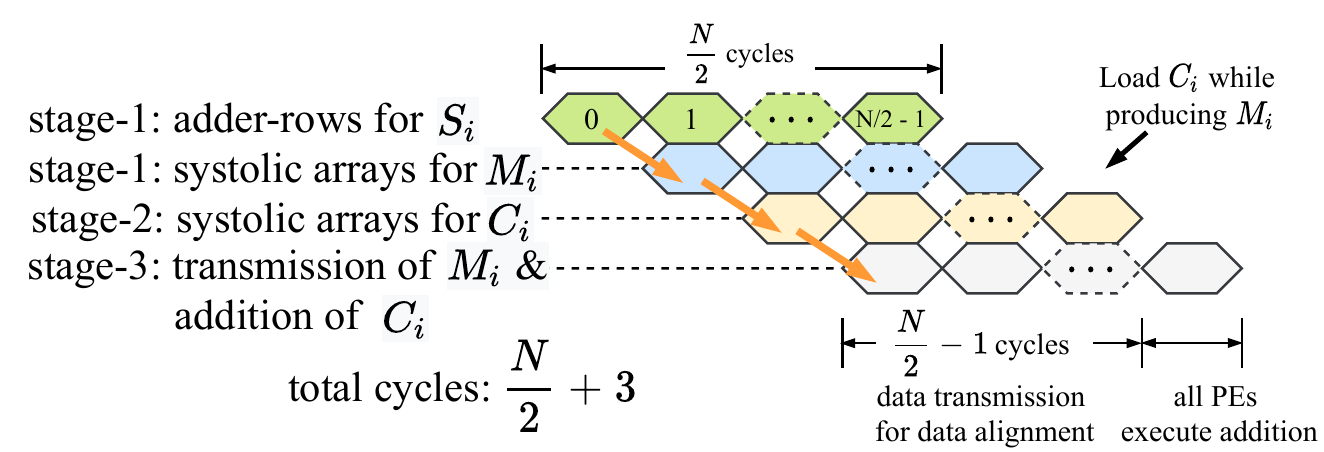}
    \caption{A simulated cycle-level illustration of the execution in a systolic tile of the Strassen's Algorithm. All 4 systolic tiles execute in parallel} 
    \label{fig:strassen_cycles}
\end{figure}

We rearrange the equations of \textit{Strassen's algorithm} into 6 groups, as shown in Equation (\ref{eq:strassen_algo}). Equations inside each group are performed in parallel, and the right-hand side of some equations in identical groups also share their operands, e.g., $A_0$, $B_0$ in the right-hand side of the upper-left equation. To harness this element-wise parallelism, the three groups in the left column can be executed in pipeline mode, the same for the three groups in the right column. For example, after calculating $M_3$, we can directly apply it to computing $C_2$, and simultaneously forward $M_3$ to the next group and calculate $C_0$ without saving or fetching the intermediate result.

With the above analysis, we group the tiles of the systolic array into clusters (or meshes). Fig.\ref{fig:strassen_scenario}(a) illustrates the layout of a systolic array cluster. Unlike the mesh architecture of \textit{Gemmini} \cite{genc2019gemmini}, data across rows and columns are also shared in diagonal directions. The shared buffers act as interfaces to the scratchpad memory. Our systolic method minimizes the I/O cost by allowing each row or column of matrix operands to enter the processing element array only once for all its associated matrix computations \cite{kung2018adaptive}. Since matrix additions are element-wise operations, addition along rows and columns are independent. Further, the matrix additions can overlap with matrix multiplications as the pipeline proceeds. In our case, every systolic array receives the result of row or column addition from a pair of 1-D adder arrays. Each 1-D adder array is either a column or a row of adders located at one side of a systolic array, as shown in Fig.\ref{fig:strassen_scenario} (a). The systolic arrays can forward the data through interconnections between neighboring systolic arrays within the ring structure.

\begin{algorithm} 
    \DontPrintSemicolon
    \SetKwInOut{Input}{Input}
    \Input{
        $a_{col}$: column index of $A[row, col]$ if sparse; \\
        $a_{val}$: input value of dense or sparse matrix; \\
        $b_{row}$: row index counter of dense $B[row, col]$; \\
        $b_{val}$:input value of dense $B[row, col]$; \\
        $a\_sparse, a\_dense$: $A$ is sparse or not; \\
        $direct\_aggr$: direct or weighted aggregation; \\
        $\mathbf{FA}$: circular $FIFO\_CAM$ (increasing order) \\
    }
    \BlankLine
    $found \leftarrow \mathbf{false}$ \;
    \eIf{ $a\_dense$ \textbf{or} $(a\_sparse$ \textbf{and} $a_{col} = b_{row})$ }{
        $a_{buf} \leftarrow a_{val}; \, b_{buf} \leftarrow b_{val}$; $found \leftarrow \mathbf{true}$ \;
        \If{$a\_sparse$}{
            $purge(\mathbf{FA})$ \;
        }
    }{
        \If{ $a_{col} > b_{row}$ } {
            $(found, a_{buf}) \leftarrow $ \tikzmarkin[set fill color=orange!50!lime!30]{sa}
                $\mathit{Find\&Skip}\left(\mathbf{FA}, b_{row}\right)$
            \tikzmarkend{sa} \;
            $b_{buf} \leftarrow b_{val}$ ; $push(\mathbf{FA}, a_{col}, a_{val})$ \;
        }
    }
    \eIf{ $found$ }{
        \eIf{$direct\_aggr$}{
            \tcc*[r]{reduction operation can be add, min, max, or mean}
            $c_{val} \leftarrow reduction\_operation( c_{val} , b_{buf} ) $ \;
        }{
            $c_{val} \leftarrow c_{val} + a_{buf} \times b_{buf}$ \;
        }
        \If{$a\_sparse$}{
            $(c_{h\_idx}, c_{v\_idx}) \leftarrow (a_{row}, b_{col})$ \tcc*[r]{indices used for write-out phase when sparse}
        }
        $b_{row} \leftarrow b_{row} + 1$; 
        $found \leftarrow \mathbf{false}$ \;
    }{
        $c_{val} \leftarrow c_{val}$ \;
    }
    $a_{out} \leftarrow (a_{val}, a_{col}); b_{out} \leftarrow b_{val}$ \;
    \caption{One cycle in each PE of the systolic array for hybrid mode matrix multiplication.}
    \label{algo:sp_mac}
\end{algorithm}

\begin{algorithm} 
    \DontPrintSemicolon
    \SetKwInOut{Input}{Input}
    \SetKwInOut{Output}{Output}
    \Input{
        \textit{\textbf{FI}}: circular FIFO\_CAM of nonzero indices; \\
        \textit{\textbf{FV}}: circular FIFO of nonzero values; \\
        $idx$: an index value to be found; \\
    }
    \Output{
        $val$: the value of nonzero found; \\
        $found$: indicates existence of $idx$; \\ 
    }
    \BlankLine
    $mask[\,\text{:}\,] \leftarrow 1$ \\
    \ForAll{ $i < \textbf{FI}.size$ }{
        \If{ $\textbf{FI}[i].idx < idx$ \textbf{ or } $i < \textbf{FI}.head$ 
        }{
            $mask[i] \leftarrow 0$ \;
        }
    }
    \tcc*[l]{ use the \textbf{leading zero detector} to find the position of first bit-one. And set the new $head$ of FIFO}
    $\textbf{FI}.head \leftarrow LeadingZeros(mask)$ \; 
    $val \leftarrow \textbf{FV}[\textbf{FI}.head]$ \;
    $found \leftarrow equal(idx, \textbf{FI}.head)$ \;
    \caption{\textbf{ Find\&Skip }}
    \label{algo:skip}
\end{algorithm}

Fig.\ref{fig:strassen_scenario} (b-d) demonstrates the execution of the right three groups in Equation (\ref{eq:strassen_algo}). The left three groups can be performed similarly and pipelined with the right three groups. Within each cycle, every 1-D adder array imports data from shared buffers to produce a row or column, and feed them to systolic array for matrix multiplication. 1-D adder arrays and systolic arrays orchestrate in a pipeline mode. Once the marginal PEs of systolic array receives the resultant row and column, they perform the MAC (Multiply-Accumulate) and forward the inputs to their neighboring PEs to perform MAC of $M_i$s, as shown in Fig.\ref{fig:strassen_scenario} (b). The MAC starts execution one cycle after the adder array as the row of Fig.\ref{fig:strassen_cycles}. As the results of matrix multiplication $M_i$ are stored in the local register of PEs (in output stationary mode), we want to reuse them for the following matrix additions, then they are transferred to the neighboring systolic array in the ring, as shown in Fig.\ref{fig:strassen_scenario} (c). As the two $C_i = C_i + M_j$ operations in Fig.\ref{fig:strassen_scenario} (b) and (c) are independent, they can be pipelined along with data forwarding of $M_i$s, but the one in stage 3 need to transmit $M_i$ to be align with those at the same coordinates in the systolic arrays prior to performing addition. All PEs perform addition simultaneously during the last cycle, as shown in the 4th row of Fig.\ref{fig:strassen_cycles}.
\begin{figure*} 
    \centering
    \includegraphics[width=0.7\textwidth]{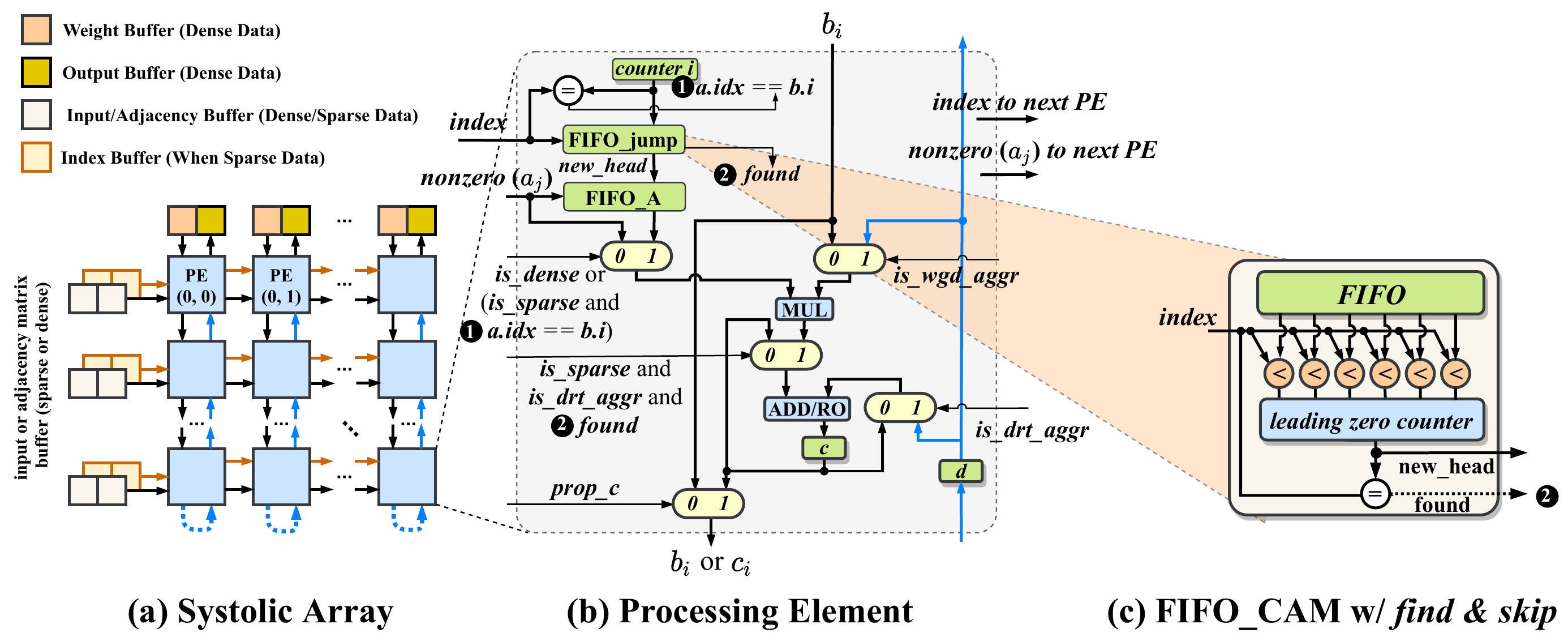}
    \caption{Layout and Logic view}
    \label{fig:pe_arch}
\end{figure*}

\begin{figure*}[h!]
    \centering
    \includegraphics[width=\textwidth]{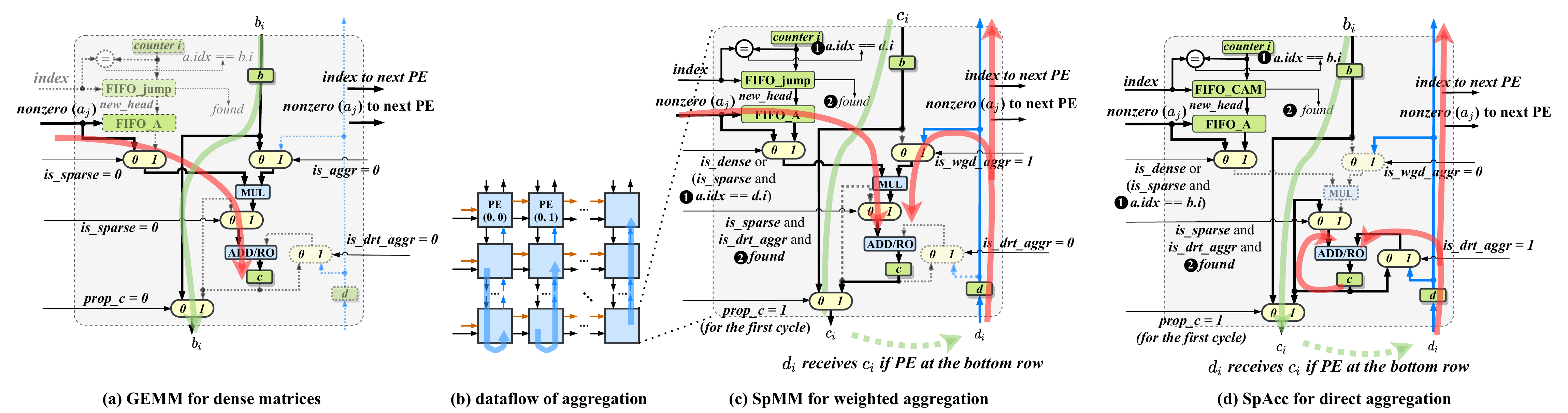}
    \caption{Three operation modes of PE: (a) dataflow of dense-dense matrix multiplication for \textit{Transformation} phase; (b) dataflow for the SpMM of weighted \textit{Aggregation}; (c) dataflow of the direct \textit{Aggregation}; where $RO$ is \textit{reduction operation}, e.g., add, min, max, etc.}
    \label{fig:sparse_pe}
\end{figure*}

With this hardware-level implementation of \textit{Strassen's algorithm}, we decrease the volume of multiplication and data transfer of operands from memory by profiting the internal bandwidth among PEs, as well as reading and writing of intermediate results from and to the scratchpad memory. \textit{Strassen's algorithm} takes an asymptotic complexity of $O(N^{2.8074})$ when applied in recursive manner, compared with $O(N^3)$ of standard matrix multiplication. As shown in Fig.\ref{fig:strassen_scenario} (b), the four tiles of systolic array execute in parallel and form a cluster, with skid buffers serving as bridges for inter-tile communication and the interface to local memory. Compared with the case in which four tiles are consolidated into one large systolic array, our design adopts a data source at the geometric center of the tile cluster, which effectively halved the data transmission path. Meanwhile, The hardware approach in \textit{VersaGNN} is in 1-level Strassen's, and the software support of 2-level Strassen's can be combined with this design for further performance gain. \cite{huang-strassen}. 


\subsubsection{Hybrid Mode Processing Elements}

Fig.\ref{fig:sparse_pe}(b) shows the internal structure of the processing element. To enable high-throughput operations on both dense and sparse matrix data structures, we devise an efficient hybrid mode \textit{processing element} (PE) of the systolic array, it equips with dedicated searchable FIFOs, with which we call it \textit{FIFO\_CAM}. The \textit{FIFO\_CAM} can search a target element within its elements and skip unused elements as depicted in Algorithm \ref{algo:skip} and Fig.\ref{fig:sparse_pe} (c). Instead of using a shared storage structure, we leverage distributed \textit{FIFO\_CAM}s design. Our design accepts the COO (Coordinate list) and CSR/CSC (Compressed Sparse Row/Column) formats \cite{wiki:sparse}. The row is interpreted as a graph node and the column indices in such row are the node's neighbors. As the column indices moving in and out of the \textit{FIFO\_CAM}, it only needs to keep a relatively small sliding window for the indices under processing. Thus, it is not necessary to store a whole list of neighbors for a graph node in the \textit{FIFO\_CAM} of a PE. Empirical result shows that 4 entries of \textit{FIFO\_CAM} is big enough to accommodate ongoing data. 

\begin{figure*}
    \centering
    \includegraphics[width=0.9\textwidth]{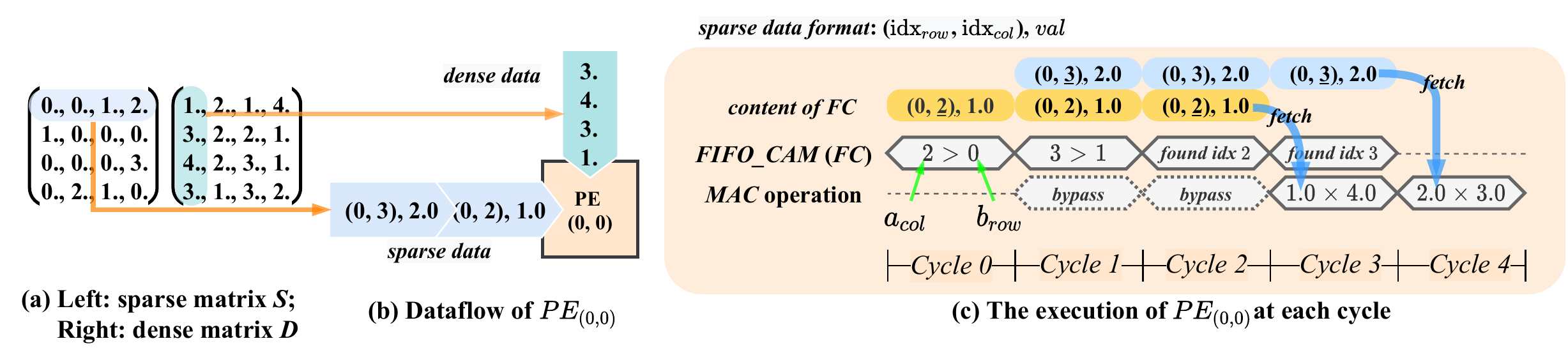} 
    \caption{An internal view on the sparse MAC of Processing Element}
    \label{fig:spmm}
\end{figure*}

As described in previous section, \textit{Transformation} phase of intermediate layers of GNNs is simply a dense-dense matrix multiplication. In Algorithm \ref{algo:sp_mac}, the PE conduct directly the MAC operation for the dense matrix multiplication, the dataflow is shown as the light green and red curve lines in Fig.\ref{fig:spmm} (a), the FIFOs for operating sparse data are bypassed and concealed. 

Recall the general mathematical expression of the convolution layer of GNN: 
\begin{equation}
    \underbrace{X^{'} = XW}_{\text{Transform}};\quad \underbrace{X^{''} = A X^{'}}_{Aggregate}
\end{equation}
In order to reuse $X^{'}$, the result of dense matrix multiplication of \textit{Transformation} phase, and to avoid transferring $X^{'}$ back to scratchpad memory, we introduce an additional dataflow path from bottom to up in PE and stored locally into register $d$ of PE, as shown in Fig.\ref{fig:pe_arch} (b). At the initial cycle of \textit{Aggregation}, the PE selects the value of $c$ register by setting signal $prop\_c$ to 1 and passes it down to the south neighbor PE. After this cycle, each PE sets $prop\_c$ back to 0 and transfer value of $c$ register as $b$ of dense matrix. At bottom row of systolic array, PEs feed back the $c$ to $d$ register of PEs, the matrix $X^{'}$ turns back from bottom to up inside the systolic array, as shown in Fig.\ref{fig:sparse_pe} (b) and (c). This manner avoids the long distance data transfer leaping across all rows used in the ring structure of \textit{RER} in \textit{EnGN} \cite{he2020engn} which leads to imbalance of data transfer rate between rows of processing array.

For the \textit{weighed aggregations}, we treat them as \textit{SpMM} ($AX^{'}$). Instead of using $b$ register of PE, now the dataflow uses $d$ register of PE, which flows into PE in opposite direction, as dense matrix element for MAC operation. The sparse data (from matrix $A$) are fed into PEs, in horizontal direction, from west to east. As described in Algorithm \ref{algo:sp_mac} and shown in Fig.\ref{fig:pe_arch}(b), PE possesses a counter, $b_{row}$, for the row number of dense matrix, it augments itself at each cycle. When condition $a_{col} = b_{row}$ satisfies, the PE performs the MAC directly as the dense-dense case. However, if condition $a_{col} > b_{row}$ meets, the \textbf{PE} checks whether \textit{FIFO\_CAM} of $A$ contains indices smaller than or equal to $b_{row}$, if there exists $a_{col} = b_{row}$, it fetches the corresponding nonzero value $a_{val}$ from \textit{FIFO} of nonzero, performs the MAC operation, and puts $a_{col}$ and the corresponding nonzero into \textit{FIFO\_CAM} and \textit{FIFO}, respectively. All i$a_{col} < b_{row}$ and their corresponding nonzeros are expelled from \textit{FIFO}'s for sparse $A$. Note that indices are enqueued into the \textit{FIFO\_CAM}s in increasing order, thus are already in sorted order in \textit{FIFO\_CAM}s. Algorithm \ref{algo:skip} describes the logic that performs searching and skipping mechanism, and we can integrate \textit{FIFO} with such logic into our \textit{FIFO\_CAM}, as shown in Fig.\ref{fig:sparse_pe}(c). The detection in \textit{FIFO\_CAM} is performed in parallel with all elements as Fig.\ref{fig:sparse_pe}(c). With the help of these \textit{FIFO\_CAM}s, the \textbf{PE} can produce one result per cycle without any inter-cycle stalls.


Fig.\ref{fig:spmm} delivers a concrete example. Suppose that  in Fig.\ref{fig:spmm}(a), $S$ and $D$ are a sparse matrix and a dense matrix, respectively. The dataflow traversing $PE_{(0, 0)}$ are the first row of $S$ in compressed format and first column of $D$ that enter the PE from left and above in Fig.\ref{fig:spmm}(b), respectively. Fig.\ref{fig:spmm}(c) demonstrates the cycle-by-cycle execution. In the first two cycles, the comparison operator does not find the matching index of current element. Thus, the \textit{FIFO\_CAM} stores them, and the \textit{MAC} bypasses the data of dense matrix to neighboring \textit{PE}s. At $Cycle 2$, the \textit{FIFO\_CAM} performs parallel comparisons of row indices of $S$ it stored with the incoming column index of $D$, and fetch the nonzero value of matched entry. Since the \textit{FIFO\_CAM} and $MAC$ are fully pipelined, at \textit{Cycle 3}, \textit{MAC} performs the multiplication and accumulation on the data fetched from previous cycle, and \textit{FIFO\_CAM} performs the comparison of new incoming index in parallel. The \textit{MAC} utilization rate relates to the number of matched indices. This issue can be solved with the algorithm introduced in Section \ref{sec:greedy}.

The direct \textit{Aggregation} can be seen as a special case of weighted \textit{Aggregation}, where every nonzero is value one. Therefore, it is useless to perform multiplication with value one. As the light-red dataflow shown in Fig.\ref{fig:sparse_pe} (c), when indices $a_{col}$ and $b_{row}$ matches each other or it found a column index in \textit{FIFO\_CAM} equal to $b_{row}$, PE directly performs the addition of $c$ and $d$ register of PE without touching multiplier.
\section{Software Approach}
\begin{figure}[h]
    \centering
    \includegraphics[width=0.4\textwidth]{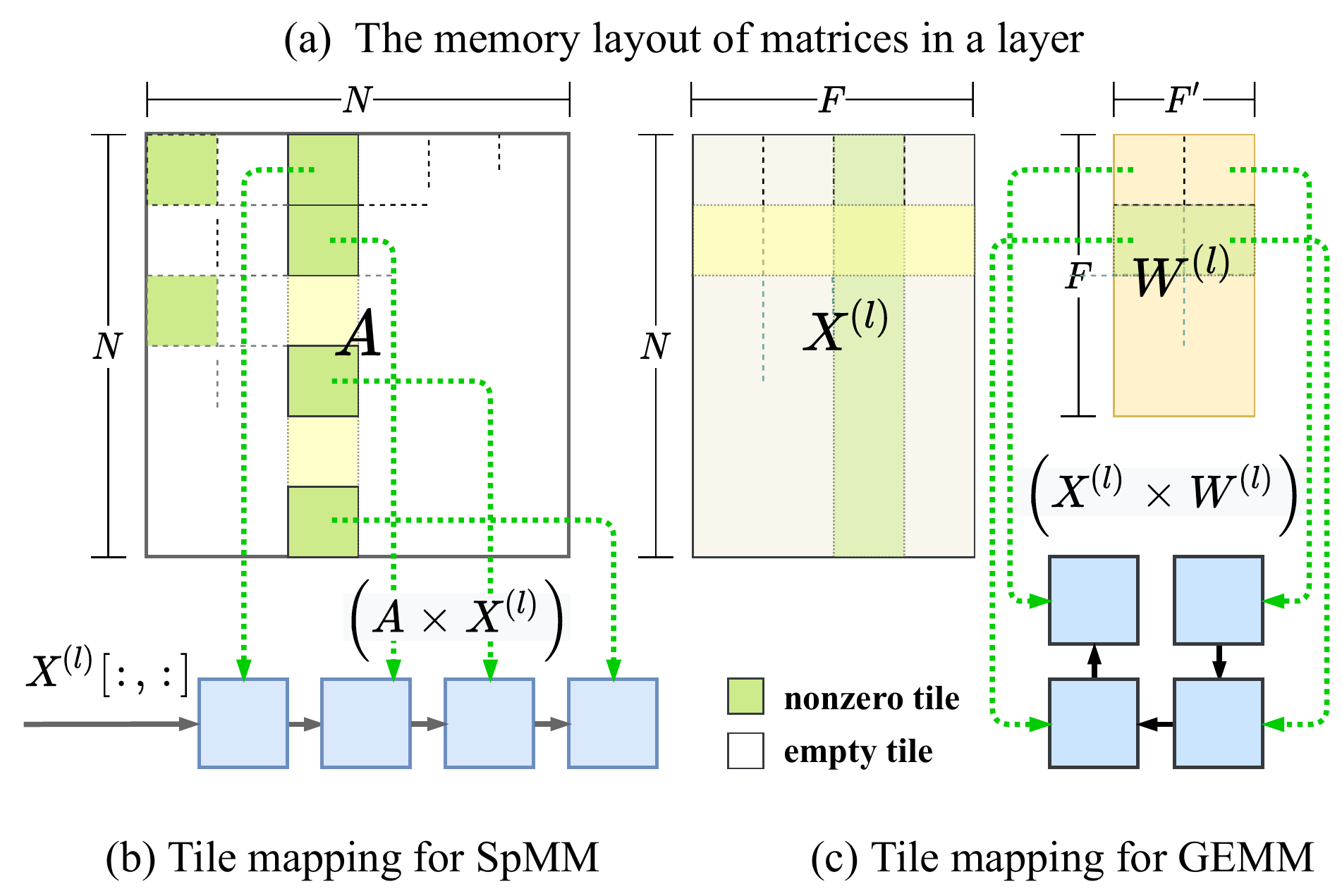}
    \caption{Memory layout and tiling strategy}
    \label{fig:tiling}
\end{figure}
\subsection{Tile Traversal strategy} \label{sec:tiling}
The vast majority of real-world graphs that GNNs operate on cannot be fitted to the limited on-chip memory of accelerators. Thus, the algorithms often divide these large-scale graphs into tiles using grid partition approaches before applying arithmetic operations, and then merge individual tiles into the full result layout.

There exist several tiling traversal strategies, including row/column-major, Z-Morton, U-Morton, and Hilbert layouts \cite{cache-oblivious, cache-1058095}, that can be harnessed by GNN accelerators. However, the shape of feature matrix $X$ and weight matrix are constantly changing through different layers. The feature matrix tends to become narrower and taller as the layers going deeper, whereas the weight matrix becomes smaller in size. As modern deep learning libraries support batched matrix multiplication with which the feature matrix can be seen as batched independent smaller matrices. In such way, small matrices are streamed into accelerator consecutively. In this work, we utilize two tile mapping strategies. For \textit{Transformation} phase, we directly map the 4 tiles in a bigger square onto the ring of systolic arrays from both input and weight matrices, as shown in Fig.\ref{fig:tiling} (c). In this way, the four systolic arrays in a ring perform the dense Strassen's algorithm as described in Section\ref{eq:strassen_algo}. While, for the SpMM of \textit{Aggregation} phase, we adopt an alternative strategy. After tiling, only tiles with nonzeros will take into account, those empty tiles are eliminated directly; the non-consecutive nonzero tiles in the same column are mapped onto the four systolic arrays, which is now in a chain, the dense tile from input matrix $X$ is then traversing the four tiles. with such manner, the systolic arrays perform the batched SpMM, as shown in Fig.\ref{fig:tiling} (b).


\begin{figure}[h]
    \centering
    \includegraphics[width=0.4\textwidth]{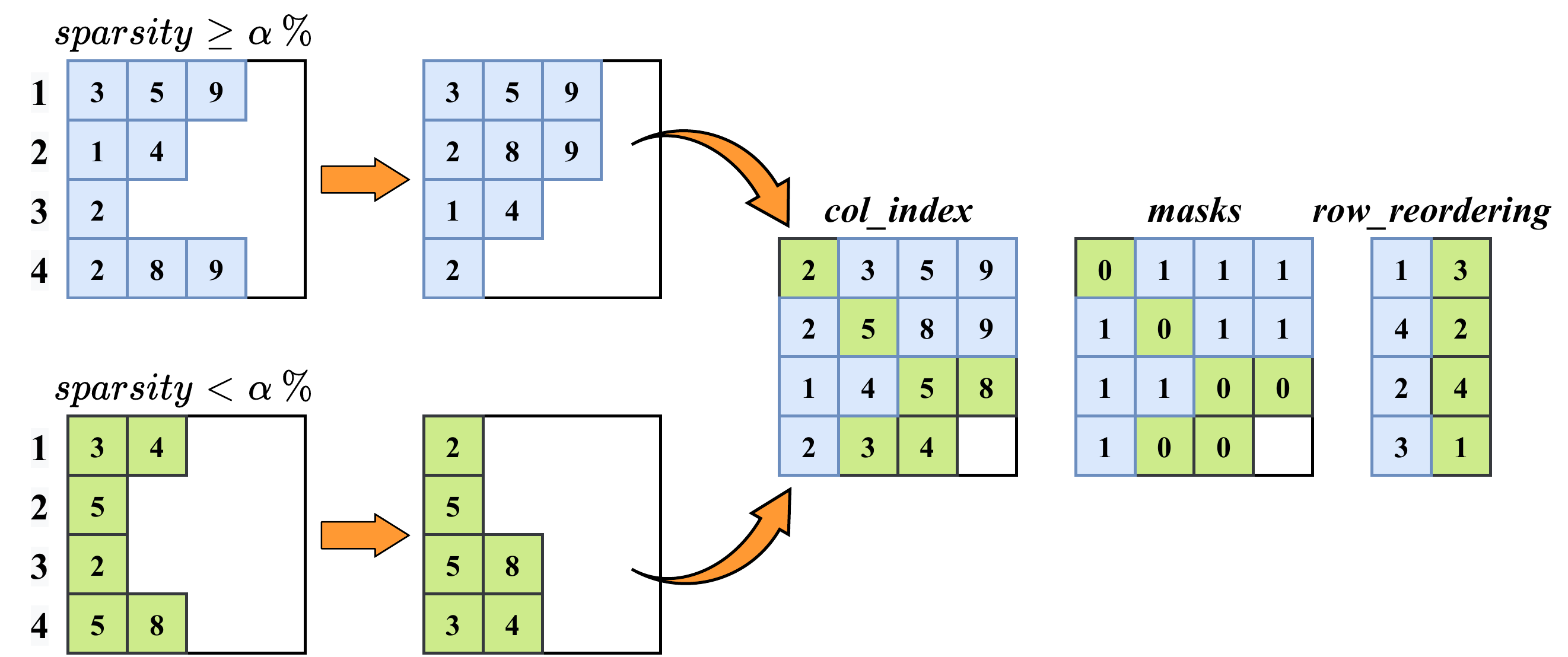}
    \caption{The greedy algorithm used for workload balancing: Given two tiles in CSR (or COO) format, each row represents one graph node, and the number in each row represent the column indices $col\_index$ (or neighboring nodes) belonging to that row (or node).}
    \label{fig:greedy}
    \vspace{-2mm}
\end{figure}
\subsection{Greedy Workload Balancing}\label{sec:greedy}
As the sparse matrix elements are irregularly distributed, some graph nodes may have relatively more neighbors. For sparse matrices in compressed formats, e.g., \textit{CSR} or \textit{COO}, each row of systolic array PEs processes features of a single node and aggregates its neighbor nodes' information. PEs in different rows will be assigned different workloads. The imbalanced workload can cause significant idling of PEs, with modules having less assigned workload finishing earlier and kept idle while waiting for those with heavier workload before the advent of next data stream, which will lead to degradation of the overall system-level performance. To remedy these issues, We introduce an effective greedy algorithm for workload balancing, which is an offline software scheme that groups tiles of sparse matrices into condensed ones. Our algorithm first sets a sparsity threshold $\alpha\%$, e.g. $40\% < \alpha \leq 50\%$. Assume the sparse adjacency matrix is split into tiles, and each tile is stored in \textit{CSR} or \textit{COO} format. For each tile with sparsity greater than $\alpha\%$, the algorithm searches for a complementary tile with sparsity less than $\alpha\%$ to combine with, as shown in Fig.\ref{fig:greedy}. Note that the number in each cell represents the column index of sparse matrix but not the value of non-zeros. Rows of the two tiles are sorted in reverse orders according to their number of elements and then combined (or packed) into one single tile. As shown in Fig.\ref{fig:greedy}, the column indices are arranged in increasing order. Duplicated elements are eliminated in each row, which prevents overlapping summation of feature vectors that belong to the same neighboring node when multiple vertices share a common set of neighbors. Our approach exploits an element mask as an identifier to trace the affiliation relationship between cells and their corresponding tiles. For combination of two tiles, each entry will have a single bit. Meanwhile, we adopt two 1-D arrays, i.e. two \textit{reorder vectors}, to record the original row ordering of tile elements. When this combined tile is fed into the systolic array, almost every PE is utilized to its full capacity during each cycle. The mask and reorder vectors are utilized by the \textit{Reordering Module} in the accelerator to direct the final results back to the scratchpad memory.
Each entry of the combined tile, in form of (($row, col$), $val$), is fetched with:
\begin{equation}
    \begin{split}
        row &\leftarrow row\_reordering[i, masks[i, j]] \\ 
        col &\leftarrow b_{col} \\ 
        val &\leftarrow values[i, j]
    \end{split}
\end{equation}
The execution of \textit{Reordering Module} can be coordinated with that systolic arrays whenever the output is ready, causing execution overhead that is generally negligible.  Further, to facilitate packed sparse tiles, both \textit{CSR} and \textit{COO} formats are treated internally as \textit{COO} within systolic arrays. This approach is extensible to combine 3 or more tiles.

\section{Evaluation}

In this section, we begin with the experimental datasets and hardware configurations. Then, we deliver the detailed analysis of our optimizations.

\begin{table*}[h]
\centering
\caption{Configurations of system}
\resizebox{0.9\textwidth}{!}{
\begin{tabular}{cccccc}
\hline
 & \textbf{PyG-CPU} & \textbf{PyG-GPU} & \textbf{HyGCN} & \textbf{EnGN} & \textbf{VersaGNN} \\
\hline
Compute Unit        & \makecell{3.0GHz @ \\ 65 cores} & \makecell{1.25GHz @ \\ 5120 cores} & \makecell{1GHz @ 32 SIMD 16 cores \\ and 32$\times$128 arrays}  & \makecell{1GHz @ 128$\times$16 arrays \\ 32 PE units in VPU}                 & \makecell{1GHz @ 8 tiles of \\ 32$\times$32 arrays}            \\
On-Chip Memory & 60MiB  & 34MiB  & 22MiB + 128KiB           & 1600KiB & 4MiB + 256KiB  \\
Peak Performance (GOP/s) & -     & -          & 8704        & 6144    & 8192  \\
Area ($mm^2$)   & -      & -     & 7.8 (12nm) & 4.54 (14nm) & 4.78 (16nm) \\
Power (W)       & 150    & 120   & 6.7       & 2.56   & 3.58 \\
Energy Efficiency (GOPS/W) & -   & -  & 1.30   & 2.4   & 1.71 \\
Area Efficiency (GOPS/$mm^2$) & -   & -  & 1.16   & 1.35   & 1.65 \\ \hline
\end{tabular}
}
\label{tab:configs}
\end{table*}

\subsection{Experiment Configurations}
\textbf{Methodology} We implemented our accelerator, \textit{VersaGNN}, along with the baselines using Chisel3 Hardware Design Language (HDL) \cite{bachrach2012chisel} Our design is also inspired by \textit{Gemmini} \cite{genc2019gemmini} and \textit{HardFloat} \cite{hardfloat}. The systolic array adopts the output-stationary fashion with 16-bit floating point input and 32-bit floating point output. We evaluated the performance of the entire system and individual modules of \textit{VersaGNN} with \textit{FireSim} \cite{firesim-isca18}, a highly efficient, open-source simulator that simulates ASIC RTL designs with timing-accurate system components, which is 
of several magnitudes faster than software-based RTL simulation. We used \textit{FireSim} to facilitate the full-system simulation by enabling integration of the simulated SoC with accurate peripheral and system-level interface models such as DDR3 memory or High Bandwidth Memory (HBM) and a last-level-cache (LLC). We synthesized \textit{VersaGNN} using open-source \textit{Yosys} and the TSMC 16nm process technology. Power and area are evaluated using a Cadence VLSI flow with TSMC 16 nm FinFET technology libraries. The placement and routing of the physical design were performed using Innovus, and power estimation using Voltus. The accelerators aim at achieving frequency of 1 GHz. To afford the high-throughput request volume, we equip the accelerator with HBM 2.0 interface with 256GB/s bandwidth, and a 256 KiB L2 Cache and a 4MiB last level cache (LLC). The energy of HBM 2.0 is estimated with 3.9 pJ/bit as in \cite{hbm2}. The configuration of \textit{VersaGNN} and the baselines are described in Table \ref{tab:configs}. 

\begin{table} 
\centering
\caption{Dataset Statistics}

\resizebox{0.48\textwidth}{!}{ 
\begin{tabular}{cccccccc} 
\hline
\textbf{Dataset}  & \textbf{Nodes}  & \textbf{Edges}  & \textbf{Features}  & \textbf{Classes}  & \textbf{Storage}  & \textbf{Sparsity}  & \textbf{Ave. Degree }\\ 
\hline
Cora (CA)         & 2,708           & 10,556          & 1,433              & 7                 & 1.5MB             & 1.44E-03         &  4  \\
Citeseer(CR)      & 3,327           & 9,104           & 3,703              & 6                 & 47MB              & 8.22E-04         &  5 \\
Pubmed (PB)       & 19,717          & 88,648          & 500                & 3                 & 38MB              & 2.28E-04         &  6 \\
IMDB-BIN (IB)     & 2,647           & 28,624          & 136                & 2                 & 1.5MB             & 4.09E-03         & 39 \\
Reddit (RD)       & 23,296          & 114,615,892     & 602                & 41                & 972MB             & 2.11E-03         & 9 \\
Amazon (AM)       & 8.6M            & 231.6M          & 86                 & 22                & 30.4MB            & 3.13E-06         & 2 \\
COLLAB (CL)       & 12,087          & 1,446,010       & 492                & 3                 & 28MB              & 9.90E-03         & 263 \\
\hline
\end{tabular}
}
\label{tab:datasets}
\end{table}

\textbf{Baselines} We choose three distinct types of baseline architectures for performance and energy efficiency comparison, including the general-purpose processors (GPP), i.e. CPU and GPU, and two state-of-the-art GNN accelerators including \textit{HyGCN} and \textit{EnGN}. We selected the server processor, an Intel Xeon (Skylake) 6151@3.0GHz processor with 512GiB DRAM, as the CPU platform. The GPU platform is equipped with NVIDIA Tesla V100 and 32GiB HBM2. The software environment for the two platforms is PyTorch \cite{paszke2019pytorch} and PyTorch Geometric (PyG) \cite{torch_geometric}. PyG is the state-of-the-art library for geometric deep learning that provides the majority of mainstream GNN models. We denote CPU and GPU platforms running PyG as PyG-CPU and PyG-GPU, respectively. The configuration of HyGCN and EnGN are listed in Table \ref{tab:configs}.

\textbf{Benchmark Graph Datasets}. Table \ref{tab:datasets} shows the statistics of benchmark graph datasets. The \textit{Feature} column specifies the length of initial feature vector ($H_0$ from Table \ref{tab:notations}). The \textit{Class} column marks the number of labels. The graphs in all listed datasets do not contain edge attributes. The sparsity of adjacency matrix is determined by the ratio of the number of graph edges to the square of the number of graph nodes. As we have stated in previous section, the \textit{Aggregation} phase, i.e. $A \times X^{'}$, is 
essentially \textbf{SpMM}, so we focus on how the sparsity affects the efficiency of the accelerator in executing SpMM. Lengths of node feature vectors determine tiling sizes and strategies of the dense matrix multiplication in the \textit{Transformation} phase.

\textbf{GNN models} We benchmark the performance of \textit{VersaGNN} using 4 GNN models, including \textit{GCN} \cite{gcn-kipf2017}, \textit{GraphSAGE (GSA)} \cite{graphsage_2017}, \textit{GIN} \cite{gin_xu2018}, and \textit{GAT} \cite{gat_2018}. The first three models are mainly used for semi-supervised classification, while \textit{GAT} can also be applied to inductive tasks, such as graph edge prediction and node feature prediction.
To make the \textit{GAT} profiting the sparse matrix multiplication and addition, we reformulate the calculation of the attention coefficients of Equation \ref{eq:gat1} and \ref{eq:gat2}. Equation \ref{eq:gat1} can be expressed in matrix form, 
\begin{equation}
    H^{'} = \mathcal{A}HW
    \label{eq:gat3}
\end{equation}
where $\mathcal{A}$ is the attention matrix and $\mathcal{A}[i, j]$ corresponds to $\alpha_{i,j}$ of Equation \ref{eq:gat1}. Suppose trainable vector $\mathbf{a}$ of Equation \ref{eq:gat2} can be split into two sections:
\begin{equation}
    \centering
    \mathbf{a}^{T}[h_i^{'} || h_j^{'}] = (\mathbf{a}_1|| \mathbf{a}_2 )^{T}[h_i^{'} || h_j^{'}]= \mathbf{a}_1^T \cdot h_i^{'} + \mathbf{a}_2^T \cdot h_j^{'}
\end{equation}
where $h_i^{'} = \Theta h_i$ and $h_j^{'} = \Theta h_j$. Turning it into matrix form, we get $H_1^{'} = H \cdot \mathbf{a}_1, H_2^{'} = H \cdot \mathbf{a}_2$. And the calculation of attention coefficient matrix $\mathcal{A}$ becomes:
\begin{equation}
    \mathcal{A} = Softmax \left( \sigma\left( Diag(H_1^{'}) \cdot A + A \cdot Diag(H_2^{'}) \right) \right)
    \label{eq:gat5}
\end{equation}
where $A$ is the adjacency matrix and $\sigma$ is the activation function \textit{LeakyReLU}. We then implement our customized code and replace the one in PyG. Most parts of Equation \ref{eq:gat3}, and \ref{eq:gat5} can be executed by SpMM, thus \textit{GAT} can also benefit from our highly efficient SpMM engine.

\textbf{Evaluation Metrics} We conduct our experiment with several metrics. We estimate 1) performance through the end-to-end inference time of GNN models; 2) throughput by billion operations per second (GOP/s); and 3) energy-efficiency by billion operations per second per Watt (GOPS/W).

\subsection{Results of experiment}

\begin{figure*}[h]
    \centering
    \includegraphics[width=0.94\textwidth]{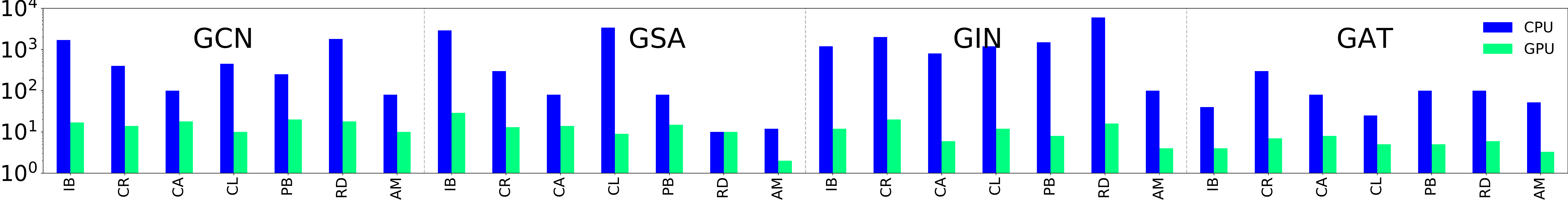}
    \caption{Performance speedup of \textit{VersaGNN} over PyG-CPU/GPU}
    \label{fig:speedup_cpu_gpu}
\end{figure*}

\begin{figure*}[h]
    \centering
    \includegraphics[width=0.94\textwidth]{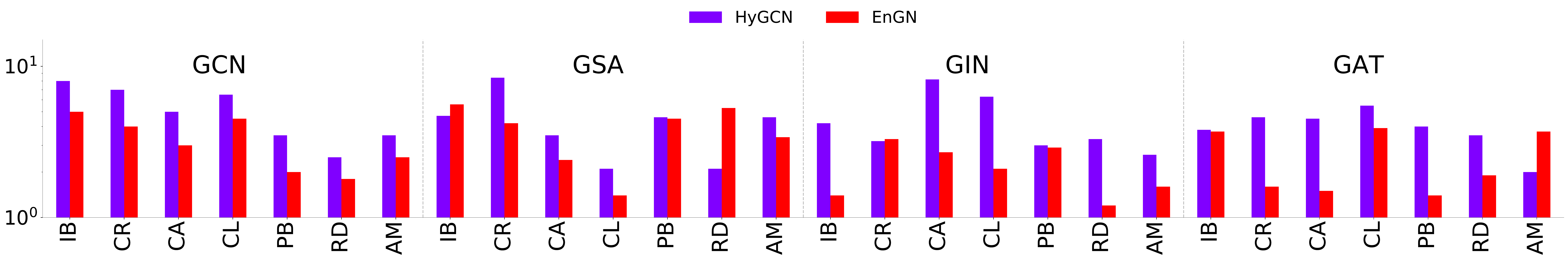}
    \caption{Performance speedup of \textit{VersaGNN} over \textit{HyGCN} and \textit{EnGN}}
    \label{fig:speedup_acc}
\end{figure*}

\begin{figure}
    \centering
    \includegraphics[width=0.4\textwidth]{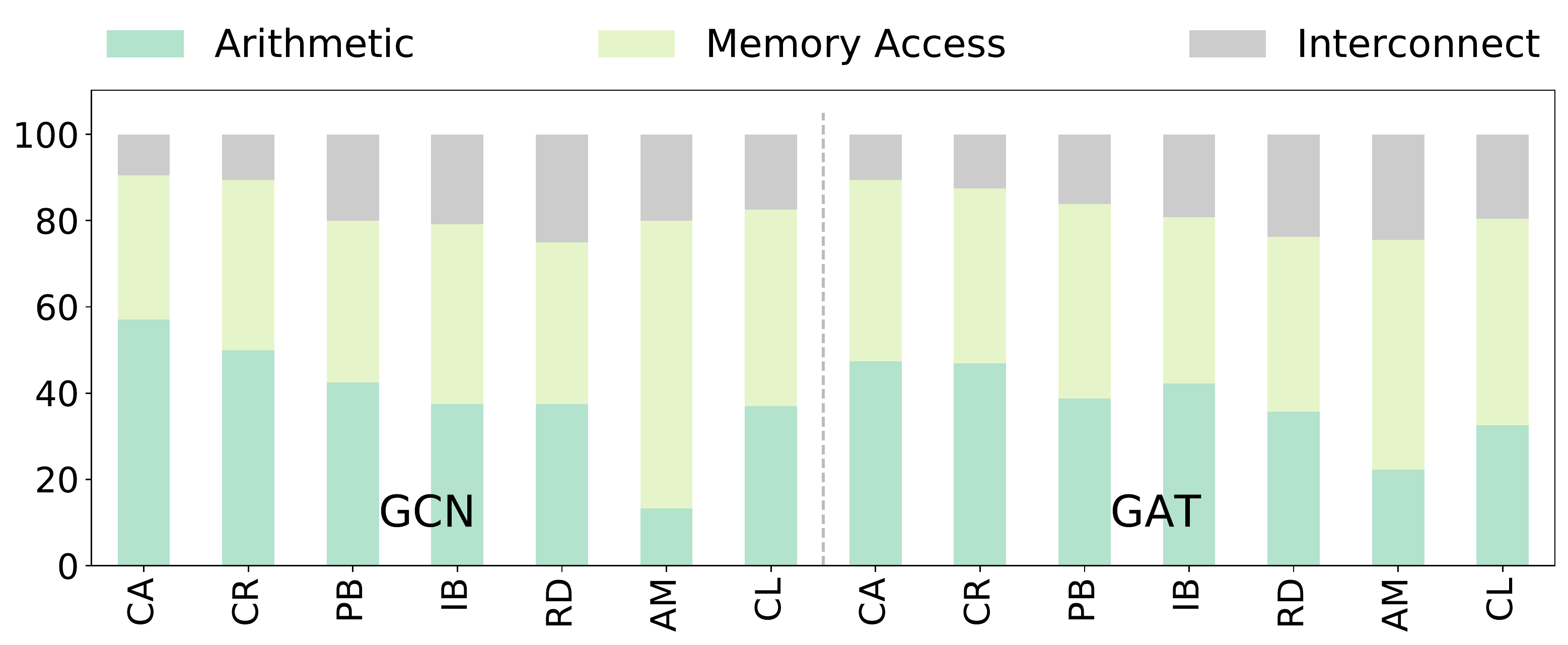}
    \caption{Energy breakdown of \textit{VersaGNN}, left: GCN; right: GAT}
    \label{fig:energy_break}
    \vspace{-2mm}
\end{figure}

\textbf{Power \& Area} We summarize the power and area of \textit{HyGCN}, \textit{EnGN}, and \textit{VersaGNN} in Table \ref{tab:configs}. As reported by the CAD tool, the 4 banked 512 KiB scratchpad memory and 128 KiB L2 Cache are the biggest part in our place-and-routed design. In the floor plan we organize the SRAMs of the accelerator in a semi-ring around the computational tiles. The second contributor of the area is the wires of the interconnections between tiles of systolic array. Due to the limit of process technology at 16nm, the power of \textit{VersaGNN} is higher than \textit{EnGN} but still achieves a higher energy efficiency than both \textit{EnGN} and \textit{HyGCN}. Static timing analysis at net-list level shows that there is still some positive slack, signifying potential for further frequency increases of our design.
Fig.\ref{fig:energy_break} provides the breakdown of the energy consumed by arithmetic operations, memory accesses, and interconnect between tiles for the model of \textit{GCN} and \textit{GAT}. As the figures illustrates that the sparser dataset tends to be compute-bound and denser dataset tends to be memory-bound. It is crucial to improve the cache utilization for the denser dataset; while sparser graph's neighbors having larger stride crossing several tiles cause cache to evict more frequently.

\textbf{Performance} The performance of \textit{VersaGNN} is compared with baseline platforms including PyG-CPU and PyG-GPU, \textit{HyGCN}, and \textit{EnGN}. The original implementation of PyG adopts the \textit{Pytorch Scatter Library} \cite{pyscatter} for the \textit{Aggregation} phase. Our experimental results show that the average performance speedup of all models on all datasets compared with PyG-CPU is 3712$\times$, as shown in Fig.\ref{fig:speedup_cpu_gpu}. For the case of GPU, we rewrite the \textit{Aggregation} function as SpMM by using \textit{PyTorch Sparse library} \cite{pysparse} and solved the memory leakage problem for the version that we used in this experiment. As the writing of this work, PyG has announced the re-implementation of the \textit{Aggregation} phase as SpMM in its future release. We obtained an average 35.4$\times$ speedup compared to PyG-GPU over all models on all datasets, as shown in Fig.\ref{fig:speedup_cpu_gpu}. Compared to \textit{HyGCN} and \textit{EnGN}, \textit{VersaGNN} achieves higher performance speedup on both small and big datasets, especially for the model with weighted \textit{Aggregation} since prior to perform the summation the neighbor node's feature vector, it needs to scale up with the coefficient, e.g., the fraction of degree term in \textit{GCN}, and the attention coefficient in \textit{GAT}. Finally, \textit{VersaGNN} is 6.32$\times$ faster than \textit{HyGCN} and 2.73$\times$ faster than \textit{EnGN}. 
\begin{figure}
    \centering
    \includegraphics[width=0.45\textwidth]{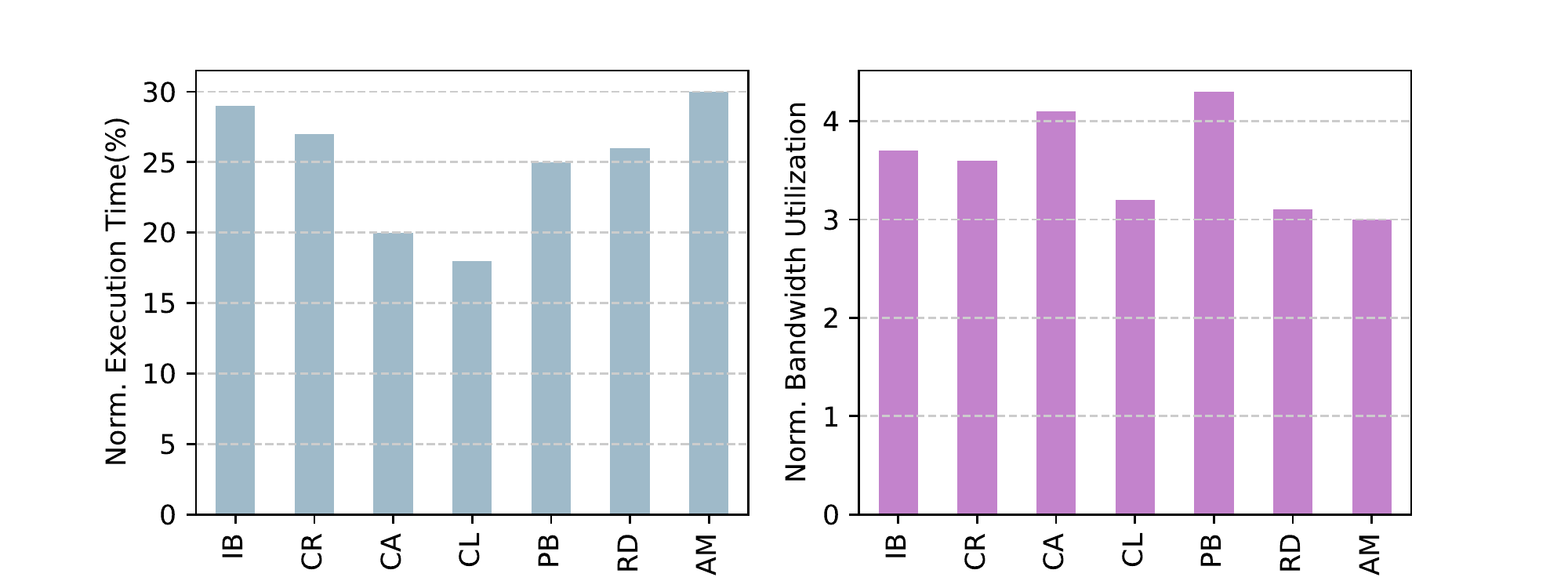}
    \caption{Effects of Strassen's algorithm for GEMM (dense) on 4 tiles of systolic array compared with a single large systolic array used by the baseline \textit{Gemmini}; Left: normalized execution time; Right: bandwidth utilization rate}
    \label{fig:bw}
    \vspace{-2mm}
\end{figure}

\textbf{Throughput} Through our experiments, we observe that datasets with higher densities of graph connections (the number of edges) or longer node feature vector tend to yield higher throughput. This is because longer feature vectors are well-suited to dense-dense matrix multiplication due to their superior memory coalescing mechanisms, whereas their high graph connectivity facilitates the memory access pattern since connected nodes are more likely to share neighbors, which increases the spatial locality of memory access. By applying greedy workload balancing, \textit{VersaGNN} is able to maintain a steady throughput even on datasets with massive irregular memory access, thus leveraging the computation and irregular memory access of the SpMM.

\subsection{Analysis of Optimization of VersaGNN}
In this section, we evaluate the performance improvement of each optimization for \textit{Transformation} and \textit{Aggregation}, respectively.

\textbf{Strassen's Algorithm} 
This optimization is applied only to the \textit{Transformation} phase which consists primarily of dense-dense matrix multiplication. The normalized execution time and the bandwidth utilization in Fig.\ref{fig:bw} show that \textit{VersaGNN} achieves 1.7$\sim$3.1$\times$ speedup when Strassen's algorithm is applied at hardware level. The performance gain is due to the parallelism from simultaneous matrix multiplications by 4 tiles of systolic array instead of one large tile, and the data transmission traversing the boundaries of neighboring tiles is achieved by utilizing the internal bandwidth of systolic arrays, which facilitates data reuse and reduces the data write-backs. This demonstrates that through collaboration, smaller spatial accelerators are able to achieve superior results than a single large module. 

\begin{figure}
    \centering
    \includegraphics[width=0.43\textwidth]{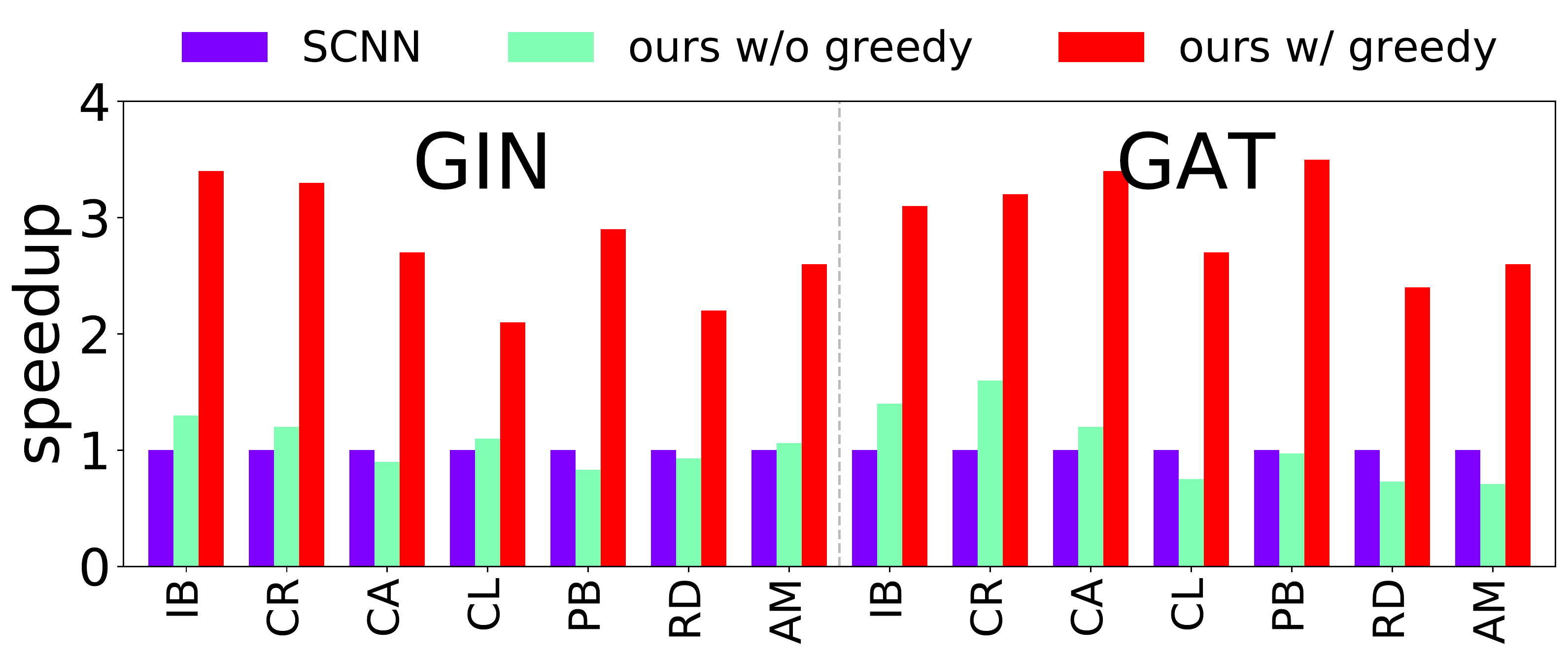}
    \caption{the speedup of SpMM of \textit{Aggregation} phase compared with SCNN \cite{scnn-2017}}
    \label{fig:speedup_scnn}
\end{figure}
\begin{figure}
    \centering
    \includegraphics[width=0.4\textwidth]{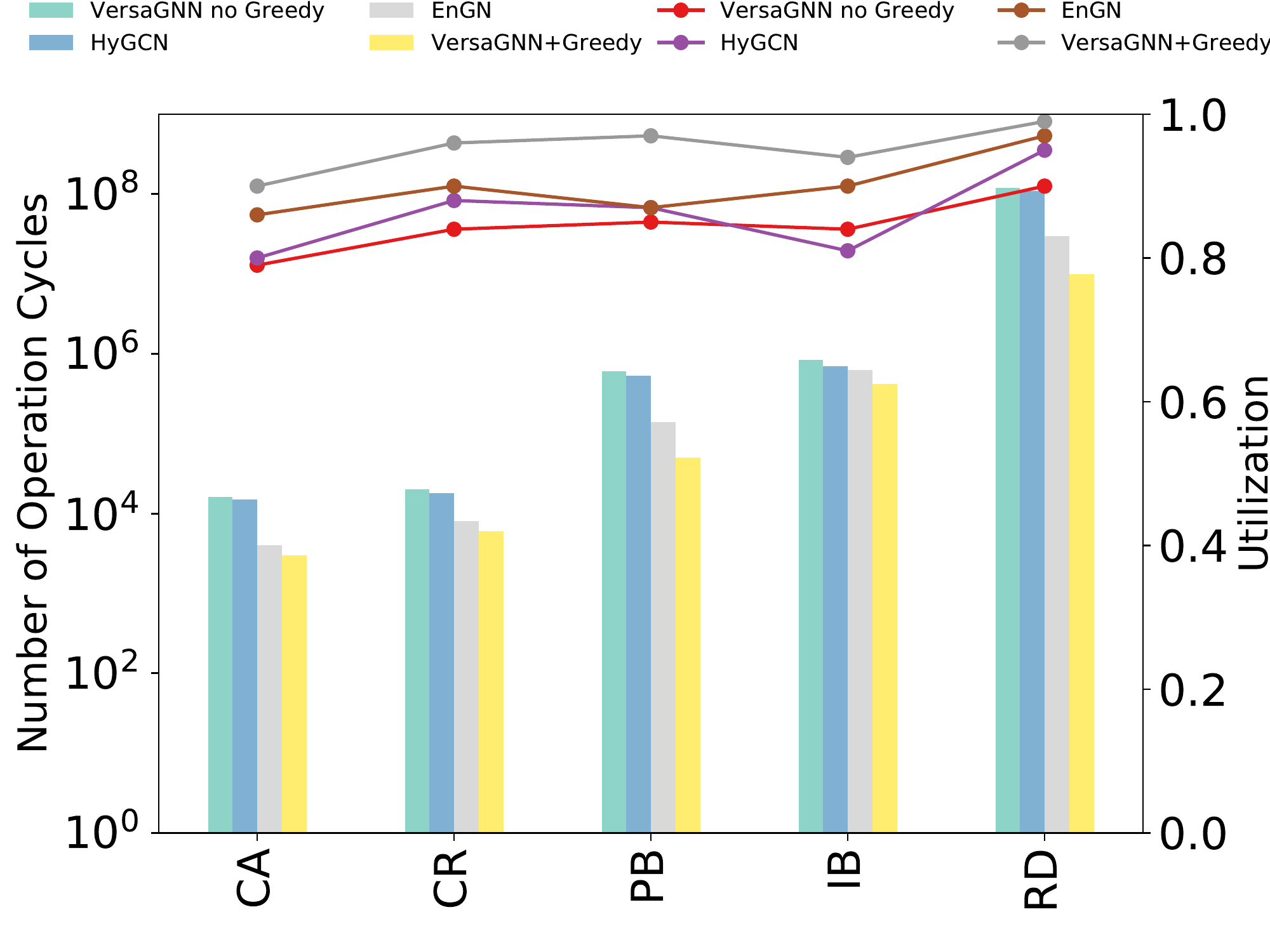}
    \caption{The utilization of PEs and overall cycles used to complete \textit{Aggregation} phase.}
    \label{fig:utilization}
\end{figure}

\textbf{Greedy Workload Balancing} The greedy algorithm introduced in Section \ref{sec:greedy} can be utilized by both direct and weighted \textit{Aggregation} phases. The efficiency of this greedy approach is also affected by the sparsity of datasets, with sparser graphs bearing more tiles to be coalesced. Graphs with higher average degree tend to produce denser tiles, and better sparse tile packing strategy delivers more parallelism and utilizes less operation cycles to improve the utilization of PEs.
To further evaluate the improvement in SpMM of \textit{Aggregation} phase, we use the \textit{SCNN}, specialized for sparse model, as baseline. Experiments show that \textit{SCNN} is less efficient for \textit{GNN}, since \textit{Cartesian Product}-based SpMM consumes most of time in processing massive irregular reduction of intermediate results, and the out-of-order scattering operation causes stall when multiple intermediate results are written into same memory location; while VersaGNN's greedy algorithm feed the operands according to the increasing order of indices in pipeline, which guarantees free of stalling, as shown in Fig.\ref{fig:speedup_scnn}.
This greedy algorithm is also affected by the tiling strategy described in Section \ref{sec:tiling}, since the sparsity varies according to the tile size and the distribution of node neighbors. Generally, the greedy workload balancing algorithm afford great improvement in performance of speed and energy saving. Since it is a static data pre-processing method prior to the execution of model, 
greedy workload balancing improves the utilization of PEs and packs the sparse tiles to increase more useful workload, as shown in Fig.\ref{fig:utilization}.
Our greedy algorithm is more flexible than the tile packing algorithm used in \textit{EnGN}, which requires two tiles with fully compatible empty slots. Moreover, \textit{HyGCN}'s window sliding strategy cannot remove zero-entries inside the tiles.
\section{Related Work}


There have been ongoing researches on tackling the hybrid computing pattern of Graph Neural Networks \cite{graphact2020, hygcn, graphzoom, auten2020hardware, geng2020awbgcn}. \textit{GraphACT} \cite{graphact2020} devises an algorithm to exploit redundant operations by looking for neighbor pairs. \textit{HyGCN} \cite{hygcn} and \textit{EnGN} \cite{he2020engn} design high-performance ASIC accelerators with two individual computing components for Aggregation and Transformation phases, respectively. \textit{GraphZoom} \cite{graphzoom} proposes an efficient clustering algorithm to condense the graph, reducing considerable inference latency.

\textbf{Deep Learning Acceleration on Sparse Structures}
The latest machine learning models, especially those for embedded and mobile systems\cite{chen2019eyeriss}, reduce their weight volumes by driving smaller weight parameters towards zeros in the feature maps and filters, leading to highly sparse models. Several works have been proposed for accelerating SparseNN and sparse matrix computing \cite{han2016eie, qin2020sigma, he2020sparse, hegde2019extensor, kung2019packing}. \cite{kung2019packing} describes a novel approach of packing sparse networks into denser formats for efficient implementation using systolic arrays. \cite{chen2019eyeriss} proposes \textit{Eyeriss v2} for execution of both compact and sparse DNNs. Architectures such as \cite{scnn-2017} and \cite{sparten} are committed to the sparse model design. Although the Cartesian-product in \cite{scnn-2017} avoids the index matching in producing products, the calculation of destination addresses corresponding to indices of products are still required while doing the partial sums, which can be seen as a postponed index matching. \textit{SparTen} \cite{sparten} provides an effective inner join mechanism, but their vector-vector multiplier cannot share broadcast inputs internally as the highly-efficient systolic array. The bandwidth is wasteful, and the broadcasting demands an intricate protocol for data synchronization. 

\textbf{Graph Analytics Systems} solve graph-related problems from different dimensions. Generally, they optimize conventional graph algorithms using a deterministic approach. Node in these graph structures typically do not possess high-dimension attributes. Therefore, the software and hardware solutions targeting these types of graphs \cite{ham2016graphicionado, challapalle2020gaas} are ineffective for GNN models. 

\section{Conclusion}
The hybrid computation mode of Graph Neural Networks impose huge obstacles in acceleration of GNN architectures. In this paper, we tackled GNN acceleration by first generalizing its computation pattern into two stages, \textit{Aggregation} and \textit{Transformation}.
\textit{Aggregation} phase is essentially formed by sparse matrix multiplication, whereas the \textit{Transformation} phase is dominated by dense matrix calculation. Then, we propose \textit{VersaGNN}, a high-throughput and memory-efficient GNN accelerator based on the systolic array paradigm. To offer the flexibility towards both dense and sparse matrix multiplication, we re-factor the processing element of systolic arrays. We further design the architecture of multiple tiles that form a computing cluster, which supports efficient execution of Strassen's algorithm. This hardware-level Strassen's algorithm dramatically reduces the computation and memory access. At the same time, We also designed a greedy workload balancing algorithm from software perspective to improve the efficiency of sparse matrix multiplication. Our vast experiments have demonstrated that, under the state-of-the-art GNN software frameworks, \textit{VersaGNN} achieves on average 3712$\times$ performance gain with 1301.25$\times$ energy reduction on CPU, and 35.4$\times$ speedup with 17.66$\times$ energy reduction on GPU.

\section{Acknowledgements}
This work is partially supported by XXX.


\bibliographystyle{IEEEtranS}
\bibliography{refs}

\end{document}